\pdfoutput=1
\documentclass[11pt]{article}

\usepackage[final]{acl}
\usepackage{times}
\usepackage{latexsym}

\usepackage[T1]{fontenc}
\usepackage[utf8]{inputenc}

\usepackage{microtype}

\usepackage{inconsolata}

\usepackage{graphicx}

\usepackage{svg}
\usepackage{mathptmx}
\usepackage{kotex}

\usepackage{booktabs}
\usepackage{tabularx}
\usepackage{multirow}
\usepackage{colortbl}
\usepackage{makecell}
\usepackage{diagbox}
\usepackage{tablefootnote}
\usepackage{threeparttable}
\title{K/DA: Automated Data Generation Pipeline for \\ Detoxifying Implicitly Offensive Language in Korean}

\author{
  \textbf{Minkyeong Jeon}\textsuperscript{1}\thanks{~~Equal contribution.},
  \textbf{Hyemin Jeong}\textsuperscript{2}\footnotemark[1],
 \textbf{Yerang Kim\textsuperscript{1}},\\
 \textbf{Jiyoung Kim\textsuperscript{3}},
 \textbf{Jae Hyeon Cho\textsuperscript{1}},
 \textbf{Byung-Jun Lee\textsuperscript{1}}
\\
 \textsuperscript{1}Korea University,
 \textsuperscript{2}Seoul National University,
 \textsuperscript{3}KAIST AI,
\\
{}
   \texttt{
   {\{alsrud1228, hs01151116, bonin147, byungjunlee\}@korea.ac.kr}}\\
   \texttt{{hyeminjeong@snu.ac.kr }}
   \texttt{{kplove01@kaist.ac.kr}}
}

\begin{document}
\maketitle
\begin{abstract}
\textbf{\textit{Caution: This paper includes content that may be considered offensive.}} \\
Language detoxification involves removing toxicity from offensive language. While a neutral-toxic paired dataset provides a straightforward approach for training detoxification models, creating such datasets presents several challenges: i) the need for human annotation to build paired data, and ii) the rapid evolution of offensive terms, rendering static datasets quickly outdated. To tackle these challenges, we introduce an automated paired data generation pipeline, called K/DA. This pipeline is designed to generate offensive language with implicit offensiveness and trend-aligned slang, making the resulting dataset suitable for detoxification model training. We demonstrate that the dataset\footnote{~Our datasets and experimental code are available at \texttt{https://github.com/minkyeongjeon/kda}.} generated by K/DA exhibits high pair consistency and greater implicit offensiveness compared to existing Korean datasets, and also demonstrates applicability to other languages. Furthermore, it enables effective training of a high-performing detoxification model with simple instruction fine-tuning. 
% We demonstrate that the dataset generated by K/DA exhibits high pair consistency and greater implicit offensiveness compared to existing Korean datasets, and can effectively train a high-performing detoxification model with simple instruction fine-tuning.
\end{abstract}

\section{Introduction}

\begin{figure}[t]
    \centering
    \includegraphics[width=0.8 \linewidth]{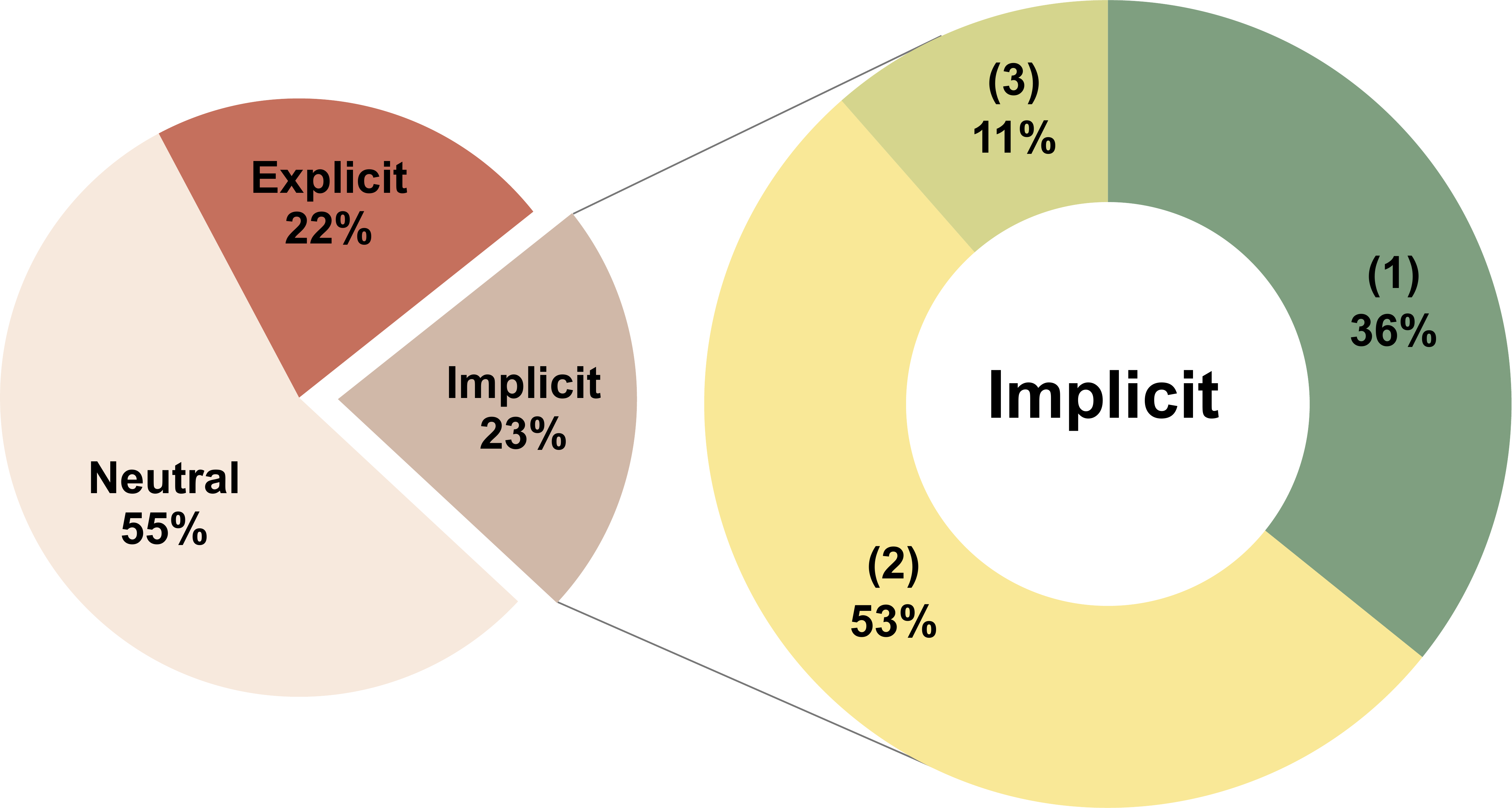}
    \vspace{0.2mm}
    {\footnotesize
    \vspace{0.8mm}
    \begin{tabularx}{\linewidth}{@{} cX @{}}
    \toprule
    \textbf{} & \multicolumn{1}{c}
    {\textbf{Examples on implicitly toxic comments}} \\
    \midrule
    \multicolumn{1}{c}{\multirow{2}{*}{(1)}} & 진지하게 가슴 작은 여자들은 \textbf{여자로 안보임.} \newline \textit{\color{gray}Honestly,  \textbf{flat-chested girls don't seem like woman}.}\\
    \midrule
    \multicolumn{1}{c}{\multirow{2}{*}{(2)}} & \textbf{김여사} 집에서 애나 보지 뭐하러 운전하냐 \newline \textcolor{gray}{\textit{Stick to babysitting, \textbf{Karen}. Who let you on the road?}}\\
    % \multicolumn{1}{c}{\multirow{3}{*}{(2)}} & \textbf{딸피} 풀악셀 사고 또 났네 \newline \textit{\color{gray}Another accident caused by a boomer with \textbf{one foot in the grave} flooring the gas pedal}\\
    \midrule
    \multicolumn{1}{c}{\multirow{2}{*}{(3)}} & \textbf{$@$}ㅐ\textbf{미} 가 너 가정교육을 어케 시킨거여\newline \textcolor{gray}{\textit{Did your \textbf{motherf*cker}} ever teach you any manners?}\\
    \bottomrule
    \end{tabularx}}
    \caption{A sample of 1,000 offensive comments was collected from Korean online communities. The implicit category was further divided into (1) disregard and mockery, (2) community-specific slang, and (3) variations of profanity and slang used to evade detection.
    }
    \label{fig:community_category}
\end{figure}

The challenge of offensive language is a continually growing concern in the field of natural language processing (NLP).
The extensive use of online social communities has created opportunities for heated discussions, which can quickly escalate to offensive or toxic levels.
%In the field of natural language processing (NLP), ethical considerations are becoming increasingly important and are being actively researched.
%many recent studies have aimed to solve ethical problems.
%Hate speech is one of the problems as profanity and slurs are causing negative social issues in various social communities.
%Here, the term \textit{slur} 
%In this paper, we specifically focus on hate speech categorized as \textit{slurs}, which are offensive words typically used to insult someone based on their identity such as race and sexuality.\footnote{https://dictionary.cambridge.org/dictionary/english/slur}
To minimize harmful social impact of offensive language, researchers have developed various datasets~\citep{hartvigsen2022toxigen, wen2023unveilingimplicittoxicitylarge, wiegand-etal-2021-implicitly-abusive} and trained detoxification models based on these datasets~\citep{dale-etal-2021-BERTdetox, Pesaranghader_2023_gpt_detox, logacheva-etal-2022-paradetox, dementieva-etal-2024-multiparadetox,lee-2020-stableTST}, aiming to purify offensiveness while retaining the original content.

\begin{table*}[!ht]
\centering
\begin{threeparttable}
    \footnotesize
    \begin{tabularx}{\textwidth}{@{} cX @{}}
        \toprule 
        & \multicolumn{1}{c}{\textbf{Examples}} \\
        \midrule
        \multirow{3}{*}{(a)} & non abuse: 구르재매맴잼잼잼잼젬 \qquad\textcolor{gray}{\textit{Supaaafunnfunfun}} \newline
        abuse: 노무현 이름뜻 \qquad\textcolor{gray}{The meaning of Roh Moo-hyun's\tnote{1} name} \newline
        not offensive: 아 중국진짴ㅋㅋㅋ 폭파시키고싶다 ㅠㅠㅠ \qquad
        \textit{\color{gray} Ah, China really LOL I want to blow it up TTT} \\
        \midrule
        
        \multirow{3}{*}{(b)} & context: <user> 좋아아아!!! \qquad
        \textit{\color{gray} coool!!!} \newline
        toxic: 너무 조장된다. 말 좀 그만하고 집 가서 혼나세요 \newline
        \textit{\color{gray} You're encouraging too much. Stop talking and go home to get scolded.} \\
        \midrule
        
        \multirow{4}{*}{(c)} & context: 저는 인종차별을 싫어합니다. 특히 흑인들이 자신의 권리를 위해 싸울 때 더욱 그렇습니다. \newline
        \textit{\color{gray}I hate racism. Especially when black people fight for their rights.} \newline
        English toxic: Only a \textbf{ni**er} would say something like that. \newline
        Translated toxic: 그런 말은 \textbf{흑인}만이 할 수 있을 거야. \qquad
        \textit{\color{gray}Only \textbf{black people} would say something like that.} \\
        \midrule

        \multirow{2}{*}{Ours} & neutral: ㅋㅋ 홈쇼핑은 \textbf{어른들} 애용하는 쇼핑몰 \qquad
        \textit{\color{gray}lol home shopping is totally an \textbf{elderly people} thing.} \\
        & toxic: ㅋㅋ 홈쇼핑은 \textbf{틀딱들}이 애용하는 쇼핑몰 \qquad \textit{\color{gray}lol home shopping is totally a \textbf{bommers} thing.}\\
        \bottomrule
    \end{tabularx}
    \begin{tablenotes}
        \footnotesize
        \item[1] The 16th President of South Korea.
    \end{tablenotes}
    \caption{Comparison of Korean offensive language datasets. (a) Human-annotated scraped dataset \citep{park2023kodoli}, containing meaningless sentences or contextually ambiguous labels. (b) LLM-generated dataset \citep{shin2023generation-komg}, producing irrelevant toxic comments to context. (c) Translated dataset \citep{song2021CADD} done by \citep{shin2022caddtranslated}, where cultural and linguistic nuances are lost. Our dataset addresses these issues by maintaining contextual coherence with challenging slurs. A detailed comparison with additional examples is provided in Appendix \ref{sec:appendix_compar}.}
    \label{tab:dataset-comparison}
\end{threeparttable}
\end{table*}

The ideal data for training a detoxification model would be a paired dataset, consisting of toxic and detoxified versions of the same content. However, a significant challenge arises from the rapid evolution of offensive language, requiring continuous scraping of online communities~\citep{park2023kodoli, jeong2022kold, lee2022kmhasmultilabelhatespeech, moon2020beepkoreancorpusonline, song2021CADD}. Without adapting to emerging offensive terms, models become vulnerable to idiosyncratic slurs~\citep{van-2018-toxic-classify-challenges}, leading to performance degradation over time. Constructing paired datasets is typically more expensive due to the need of human annotation, and involving humans in continuously updating the model to address contemporary offensive language would be prohibitively costly. Leveraging language models to generate offensive examples ~\citep{shin2023generation-komg, hartvigsen2022toxigen} can reduce this expense. However, these models struggle to generate recent offensive terms that they have not been trained on, and we found that off-the-shelf LLMs are generally weak at generating offensive language; see Table~\ref{tab:dataset-comparison} for examples from different previous methods.

Another important consideration is ensuring that there is a sufficient amount of \textit{implicitly toxic data} in the dataset. 
% This type of data may not be overtly offensive but can carry harmful or derogatory meaning in context, making it more difficult to detect and detoxify~\citep{breitfeller-etal-2019-finding, macavaney2019hate}.
This type of data may not include profanity or swear words, but still can carry derogatory meaning such as sarcasm or social bias in context, making it more difficult to detect or collect~\citep{breitfeller-etal-2019-finding, macavaney2019hate}.
It is thus crucial to include an adequate volume of implicitly toxic data to enable the model to be trained to effectively handle various forms of implicit offensiveness~\citep{wiegand-etal-2021-implicitly-abusive}. 

In particular, Korean exhibits distinct forms of mockery, sarcasm, and wordplay that are deeply tied to the nuances of the language~\citep{Yook1999crosslanguage, Merkin2009korean-}. Unlike English, which is an inflectional language, Korean is an agglutinative language and allows for a wider range of sarcastic tones through word variations. This linguistic structure makes it difficult for models trained on translated English datasets to interpret implicit expressions accurately.

Figure~\ref{fig:community_category} shows the proportion and examples of implicitly toxic texts, revealing that they exist at a similar rate to explicitly toxic content in actual online comments. However, we found that language models also tend to focus on explicit offensiveness, resulting in the issue that automatically generated data from these models contains a lower proportion of implicitly toxic texts.

To address these issues, we introduce an automated pipeline for synthesizing paired offensive language data, which we call \textbf{K}orean offensive language \textbf{D}ata generation \textbf{A}utomation (K/DA). 
The main contributions of this paper are as follows: 
1) We introduce an automated pipeline, K/DA, for generating paired synthesized dataset of neutral and toxic texts. This pipeline integrates recently emerging offensive language and ensures high-quality results by filtering out low-quality outputs. {We further demonstrate its 
scalability by applying the same pipeline to different language and model types, showing its language- and model-agnostic nature.} 2) We provide a language detoxification dataset with around 7.5K neutral-toxic pairs. Unlike previous offensive or toxic language datasets, this paired structure facilitates easier model training and encompasses a broader range of offensive language, including explicit profanity, implicit offensiveness, and their variations. 3) Our experiments show that models trained on our dataset achieve improved detoxification performance.

\section{Related Works}
\paragraph{Offensive language datasets}
\begin{figure*}
    \centering
    \includegraphics[width=1\linewidth]{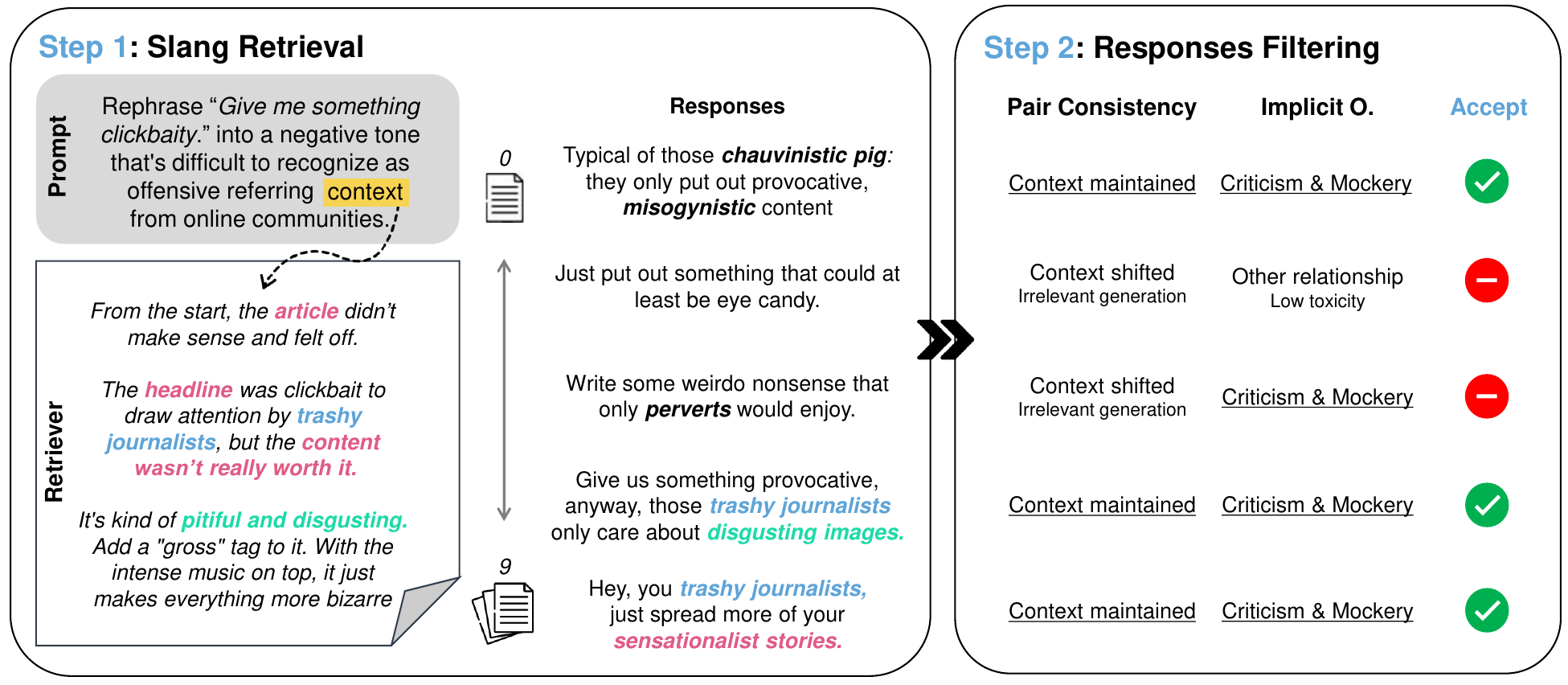}
    \caption{An overview of K/DA, the pipeline for automated offensive language data generation, is provided. In the first stage, slang retrieval, five outputs are generated using varying amounts of retrieved context. These generations are then filtered based on two key criteria: pair consistency and implicit offensiveness (Implicit O.). Only the high-quality generations that pass both filters are included in the final dataset.
    }
    \label{fig:dataset generation pipeline}
\end{figure*}

Although numerous studies have focused on creating datasets for offensive language detection~\citep{zampieri-etal-2019-predicting, davidson2017automated, song2021CADD, hartvigsen2022toxigen}, most of these resources are unfortunately limited to English. The scarcity of offensive language datasets in other languages hampers the performance of offensive language detection and detoxification systems in those languages. Previous research has also highlighted the importance of developing such datasets in other languages to address these gaps~\citep{pitenis2020offensivelanguageidentificationgreek,mubarak2020arabic,diaz-torres-etal-2020-automatic}. Translating existing datasets into the target language is a possible solution, but offensive language is highly dependent on the cultural, political, and linguistic context of its original language~\citep{koppel-ordan-2011-translationese}. As a result, translated offensive texts often lose their nuance and impact, as shown in Table \ref{tab:dataset-comparison}. The Korean language, in particular, is rooted in a distinct cultural and contextual framework compared to other languages, making it necessary to develop offensive language datasets specific to Korean. While several studies have created non-paired Korean datasets for offensive language detection using human annotation~\citep{jeong2022kold, park2023kodoli, lee2022kmhasmultilabelhatespeech, moon2020beepkoreancorpusonline} or machine generation ~\cite{shin2023generation-komg}, training detoxification models is more natural and efficient with paired datasets, which is the primary focus of our work.

\paragraph{Data generative methods}
Due to the rapid evolution of offensive language, continuously updating datasets to include new terms is essential. This poses a significant challenge for approaches that rely on human annotation~\citep{kennedy2020constructingintervalvariablesfaceted, qian2019benchmarkdatasetlearningintervene}. While methods using language models for dataset generation~\citep{hartvigsen2022toxigen, shin2023generation-komg} reduce the need for human labor, they can still encounter the same problem if the models are not updated to recognize new offensive terms. To address these challenges, this paper introduces a pipeline that leverages LLMs and Retrieval-Augmented Generation (RAG) to generate datasets aligned with real-world language trends, enabling more efficient updates without relying on human labor.

\section{Implicit Offensiveness and Trend-Aligned Slang}
Implicitly offensive language is defined as a tone of disregard or mockery to insult while avoiding explicit slurs or profanity~\citep{wiegand-etal-2021-implicitly-abusive}. This definition helps group challenging examples of offensive language that are often mishandled by trained models, allowing us to specifically target these difficult cases. Several previous studies on offensive language have addressed implicit offensiveness, such as sarcasm through rhetorical expressions~\citep{moon2020beepkoreancorpusonline} or stereotype-based rude jokes~\citep{park2023kodoli}.

However, we found that this definition does not fully capture the characteristics of real-world conversations. Figure~\ref{fig:community_category} illustrates the types of offensive comments collected from Korean online communities, categorized using GPT-4 Turbo. Upon further investigation, we were able to divide the implicitly offensive comments into three subcategories: (1) disregard and mockery, consistent with past definitions of implicit offensiveness, (2) community-specific slang that is familiar within certain groups but difficult for outsiders to interpret, and (3) variations of profanity used to avoid detection. The figure reveals that the majority (64\%) of implicitly offensive comments fall under categories (2) and (3), which have not been extensively studied in prior research. This highlights the need for a dataset containing sufficient examples of these types.
To address this, we specifically coin the term \textit{trend-aligned slang} to describe categories (2) and (3).
These newly defined forms of implicit offensiveness present a unique challenge compared to the conventional definition, as these slang are localized within specific communities and evolve rapidly. Trend-aligned slang is continuously developed through various online disputes. As community administrators attempt to censor the use of these emerging slang, they morph into variations, using phonetic or visual similarities that are easily understood by humans but are challenging for models to detect; see Appendix Table~\ref{tab:Categorization of variations} for examples.

% Given the impracticality of updating slang frequently, developing an effective method for collecting dataset becomes crucial~\citep{van-2018-toxic-classify-challenges}. Furthermore, community-specific slang includes both toxic words and hate speech targeting specific groups, making it difficult to draw a clear distinction between them. Previous research~\citep{fortuna-etal-2020-toxic} highlighted the lack of standardized labeling criteria across datasets, with some studies reclassifying hate speech as toxicity to establish a common criterion. Given this ambiguity, we construct K/DA dataset mainly targeting implicitly offensive language capturing both expressions used in real-world, instead of drawing rigid categorical boundaries. Our experimental results demonstrate that the detoxification models trained with this expanded definition not only enhances detoxification performance, also preferred to human evaluators.}

Given the impracticality of continuously updating slang, developing an effective dataset collection method is crucial~\citep{van-2018-toxic-classify-challenges}. Moreover, trend-aligned slang encompasses both toxic language and hate speech targeting specific groups, making it difficult to establish clear distinctions. Previous research~\citep{fortuna-etal-2020-toxic} pointed out the lack of standardized labeling criteria and reclassified hate speech as toxic in certain datasets to ensure consistency. Given this ambiguity, we construct the K/DA dataset to primarily capture implicitly offensive language as used in real-world contexts, rather than imposing rigid categorical boundaries.
% Experimental results show that detoxification models trained with this expanded definition achieve improved detoxification performance.

\section{K/DA: Automated Korean Offensive Language Data Generation Pipeline} 

Based on the discussions so far, our data generation pipeline must meet the following requirements:
\begin{enumerate}
    \item \textbf{Paired dataset:} A dataset containing pairs of neutral sentences and their offensive counterparts is essential for the straightforward training of detoxification models.
    \item \textbf{Trend alignment:} The pipeline should generate data incorporating recently developed trend-aligned slang to ensure that trained models remain effective over time.
    \item \textbf{High toxicity:} Simply scraping data often leads to examples that contain only neutral textual expressions, diminishing the dataset's effectiveness. The pipeline must ensure the inclusion of highly toxic content.
\end{enumerate}

To fulfill these criteria, we propose a two-stage data generation process: (1) \textit{slang retrieval} and (2) \textit{generation filtering}. In the first stage, outputs with trend-aligned slang are generated from neutral sentences,\footnote{We used \textit{Topic-Based Informal Social Media Corpus}~\citep{AIHub}.} by leveraging context retrieved from online communities. In the second stage, two filtering criteria are applied to refine the generations, ensuring they satisfy two essential factors: preserving the original context and exhibiting sufficient (and implicit) toxicity. A summary of this data generation pipeline is shown in Figure~\ref{fig:dataset generation pipeline}.

\subsection{Slang Retrieval}
To stay aligned with the rapidly changing nature of slang, it is essential to develop a dynamic data generation pipeline rather than depending on a static dataset. However, as shown in Table~\ref{tab:dataset-comparison}, previous methods have significant limitations. Naive generation from language models tends to produce less toxic content and suffers from the same issues as static datasets when the language model is not updated to reflect current trends. Moreover, simple web scraping frequently results in irrelevant or meaningless sentences, which can adversely impact the performance of models trained on this data.

To generate a paired dataset with trend-aligned, highly toxic slang, we employ Retrieval-Augmented Generation (RAG, \citealp{NEURIPS2020_RAG}). By retrieving trend-aligned slang from Korean online communities\footnote{
We used dataset scraped from a \textit{DCInside, FM Korea, and Ilbe}. These platforms are comparable to Reddit in terms of slang evolution and implicit toxicity.} and augmenting neutral sentences, we generate toxic versions that preserve the original context, forming neutral-toxic pairs for the dataset. We start by building a vector database by embedding 92,953 sentences crawled from Korean online communities using SBERT~\citep{reimers2019_sbert}. Slang relevant to the context of the neutral sentences, determined by cosine similarity from the vector database, is incorporated into the prompt to guide LLMs in generating corresponding offensive language. See Appendix~\ref{sec: slang retrieval} for detail RAG setup.

\paragraph{Multiple RAGs for maximized diversity} A well-known limitation of the conventional RAG approach, which fixes the number of retrievals $n$, is that irrelevant information may be retrieved if the vector database lacks sufficient slang relevant to the current context. Reducing $n$ to avoid irrelevant retrievals can, however, compromise the diversity of the generated outputs. \citet{asai2023selfRAG} proposed a solution by training a language model to dynamically determine $n$, but this requires additional costs for dataset preparation and model training.

To address this without the need for additional model training, we apply RAG multiple times with different values of $n$ and forward all retrieval results to the filtering stage. The filtering process removes toxic augmentations that fail to preserve context due to irrelevant retrievals. Therefore, when the filtering works effectively, this approach ensures the generated outputs maintain relevance while maximizing diversity. In K/DA, the number of retrievals $n$ is set to $\{0, 3, 5, 7, 9\}$.
% , where $0$ is used to generate offensive language solely from the prompt, without utilizing content from the vector database. 
We conduct an empirical analysis showing that retrieval with different $n$ values is crucial for maximizing the potential quality of generations before filtering. For detailed analysis, see Appendix~\ref{sec: slang retrieval}.

% 아래 부분 appendix로 이동
% Figure~\ref{fig:selfRAG} provides an empirical analysis showing the rank distribution of different $n$ values when GPT-4 Turbo is asked to evaluate the quality of retrieval results. While $n=3$ and $n=0$ tend to have higher proportions of top rankings, $n=7$ and $n=9$ also capture a significant share of the highest ranks. This suggests that retrieving with diverse $n$ values is crucial for maximizing the potential quality of generations before filtering. For detailed generation prompt for slang retrieval, see appendix, Table~\ref{tab: Prompt Template for RAG}.

\subsection{Filtering Augmented Generations}
\label{sec:Filtering Augmented Generations}

Due to either irrelevant retrievals or limitations of the LLM, slang retrieval can sometimes produce inadequate generations. We identified following three types of low-quality outputs: \textbf{(1) Answer generation}: The LLM interprets the reference neutral sentence as a question and responds to it, rather than turning it into an offensive statement. \textbf{(2) Irrelevant generation}: The LLM misinterprets the reference, producing irrelevant generations or introducing inappropriate slang. \textbf{(3) Inoffensive generation}: The LLM fails to make the sentence offensive, which frequently occurs with certain types of reference sentences, such as factual statements or information requests.
% \begin{enumerate}
%     \item \textbf{Answer generation}: The LLM interprets the reference neutral sentence as a question and responds to it, rather than turning it into an offensive statement.
%     \item \textbf{Irrelevant generation}: The LLM misinterprets the reference, producing irrelevant generations or introducing inappropriate slang.
%     \item \textbf{Inoffensive generation}: The LLM fails to make the sentence offensive, which frequently occurs with certain types of reference sentences, such as factual statements or information requests.
% \end{enumerate}
To filter out these low-quality outputs from slang retrieval, we introduce a two-stage filtering process. The first stage removes inconsistent pairs (1 and 2), while the second stage eliminates outputs with insufficient implicit offensiveness (3). This filtering is performed by the LLM itself, reducing the reliance on human labor and aligning with recent trends~\citep{chiang2023_LLM_can_be_annotator, liu2023geval}. 
    
    % Appendix로 이동
    % \begin{figure}
    %     \centering
    %     \includegraphics[width=0.91    \linewidth]{latex/figures/GPT rankings of retreivals.pdf}
    %     \caption{The rankings of retrievals of different $n$ s for slang retrieval, where GPT-4 Turbo is asked to rank the retrievals based on their implicit offensiveness. The generations are ranked from 1st to 5th, with 1st being the most aligned.}
    %     \label{fig:selfRAG}
    % \end{figure}

    \begin{table*}[!ht]
        \centering
        \begin{tabular}{@{}clcccc}
        \toprule 
            &\textbf{} & \textbf{Overall O.} & \textbf{Implicit O. ($\uparrow$)} & \textbf{Consistency ($\uparrow$)} & \textbf{Retained ($\uparrow$)} \\ 
            \midrule
             & \multicolumn{1}{c}{Unfiltered} & 2.399$_{(\pm0.058)}$ & 2.331$_{(\pm0.053)}$ & 3.804$_{(\pm0.045)}$ & 100.00 \% \\  
            \midrule
            \multicolumn{5}{l}{\textbf{Pair consistency filtering}}\\
             &Context Shift & 2.174$_{(\pm0.059)}$ & 2.092$_{(\pm0.053)}$ & \underline{4.182}$_{(\pm0.036)}$ & 47.89 \%\\ 
             &QA and Paraphrasing & 2.438$_{(\pm0.059)}$ & 2.332$_{(\pm0.053)}$ & 3.853$_{(\pm0.042)}$ & 64.20 \%\\ 
             &QA & 2.464$_{(\pm0.059)}$ & 2.403$_{(\pm0.053)}$ & 3.479$_{(\pm0.049)}$ & 64.26 \% \\ 
             & C. S. \& QA and P. & 2.043$_{(\pm0.053)}$ & 2.263$_{(\pm0.056)}$ & \textbf{4.223}$_{(\pm0.035)}$ & 34.80\% \\  
            \midrule
            \multicolumn{5}{l}{\textbf{Implicit offensiveness filtering}}\\
             &Derogatory Detection & 2.719$_{(\pm0.057)}$ & \underline{2.622}$_{(\pm0.050)}$ & 4.060$_{(\pm0.033)}$ & 63.24 \% \\
             &Tone Classification & 2.139$_{(\pm0.059)}$ & 2.166$_{(\pm0.106)}$ & 4.198$_{(\pm0.036)}$ & 18.60 \%\\ 
             &Multi-meaning Relationship & 2.127$_{(\pm0.058)}$ & \textbf{3.159}$_{(\pm0.264)}$ & 4.212$_{(\pm0.035)}$ & 3.20 \%\\  
            \bottomrule
        \end{tabular}
        \caption{G-Eval results for datasets filtered according to different filtering prompts. The retained column shows the ratio of generation retained after filtering. The numbers in parentheses indicate the standard error.}
        \label{tab: Evaluation of Filtering Method}
    \end{table*}

    \paragraph{Filtering for pair consistency}

Ensuring consistency between paired sentences, so that they convey the same meaning, is crucial for building an effective dataset to train detoxification models.
The core idea behind our approach is to introduce the LLM to the identified types of inconsistent pairs, as well as more specific subtypes, and ask whether the generated pairs fall into these categories. This includes prompting the LLM to determine if the generated output is a response, a paraphrasing, or has an arbitrary relationship to the neutral sentence. If the LLM deems the generated pair consistent, it is retained; otherwise, it is discarded. Empirically, we found that providing a one-shot example for each type of pair results in the most effective filtering. The exact structure of the prompt is shown in the Appendix Table~\ref{tab: Prompt Template and One-shot Example for Context Preservation Filtering} and Table~\ref{tab: Pair Consistency Filtering prompts and examples}.

    \paragraph{Filtering for implicit offensiveness}
When the topic of a neutral sentence is less controversial, retrievals from the vector database tend to have lower toxicity, leading to inoffensive generations. Conversely, when the topic is highly controversial, the retrievals may be filled with explicit profanities, resulting in explicitly offensive outputs rather than implicitly offensive ones. Since our goal is to create a dataset with a high proportion of implicitly offensive language, both of these scenarios need to be discarded. Similar to the filtering process for pair consistency, we provide the LLM with definitions of trend-aligned slang and implicit offensiveness, along with a few-shot examples. We then prompt the LLM to evaluate whether the generated output includes the desired trend-aligned slang and implicit offensiveness. Unlike using the LLM for direct generation of implicitly offensive language, we found this approach to be very effective in distinguishing the targeted implicitly offensive content with trend-aligned slang from other types. The complete filtering prompts can be found in the Appendix Table~\ref{tab: Prompt Template for Toxicity Filtering} and~\ref{tab: Toxicity Preservation Filtering prompts and examples}.

\section{Experiments}

\begin{table*}[!ht]
    \centering
    \begin{tabular}{ccccc}
    \toprule
        \textbf{Datasets} & \textbf{Overall O.} & \textbf{Implicit O. ($\uparrow$)} & \textbf{Consistency ($\uparrow$)} \\ \midrule
        K-OMG~\citep{shin2023generation-komg} & \textbf{3.770}$_{(\pm0.040)}$ & 2.399$_{(\pm0.054)}$ & 1.393$_{(\pm0.030)}$ \\ 
        BEEP~\citep{moon2020beepkoreancorpusonline} & 2.300$_{(\pm0.055)}$ &  2.206$_{(\pm0.048)}$ & - \\
        KODOLI~\citep{park2023kodoli} & 3.293$_{(\pm0.058)}$ & 2.554$_{(\pm0.047)}$ & - \\
        Translated CADD~\citep{shin2022caddtranslated} & 2.963$_{(\pm0.055)}$ & 1.861$_{(\pm0.053)}$ & 1.458$_{(\pm0.036)}$ \\ \midrule
        Ours & 2.719$_{(\pm0.057)}$ & \textbf{2.622}$_{(\pm0.050)}$ & \textbf{4.060}$_{(\pm0.033)}$ \\
    \bottomrule
    \end{tabular}
    \caption{\label{tab:G-Eval of other dataset}G-Eval results on 500 toxic–neutral pairs. Consistency is only computed for paired dataset. The numbers in parentheses indicate the standard error.}
\end{table*}

% \begin{table}[!ht]
%     \centering
%     \begin{tabular}{ccccc}
%     \toprule
%      & \textbf{O} & \textbf{I} & \textbf{C} & \textbf{F} \\ \midrule
%      \textbf{K-OMG} & \makecell{3.24\\\addlinespace[-1mm]$_{[0.91]}$} & - & \makecell{4.17\\\addlinespace[-1mm]$_{[0.26]}$} & \makecell{4.32\\\addlinespace[-1mm]$_{[0.61]}$} \\ \addlinespace[3mm]
%     \textbf{Ours} & \makecell{4.196\\\addlinespace[-1mm]$_{[0.924]}$} & \makecell{4.196\\\addlinespace[-1mm]$_{[0.889]}$} & \makecell{3.905\\\addlinespace[-1mm]$_{[0.804]}$} & \makecell{4.108\\\addlinespace[-1mm]$_{[0.725]}$} \\ 
%     \bottomrule
%     \end{tabular}
%     \caption{\label{tab:human eval of other dataset}Human evaluation result of 50 random samples from K/DA and K-OMG. The numbers in parentheses represent the Cronbach's $\alpha$.}}
%     \end{table}

To demonstrate the effectiveness of the proposed method, we evaluate the quality of the dataset generated using the K/DA pipeline in Section~\ref{sec:Evaluation of Dataset}. Additionally, we conduct further experiments across different languages and models to validate the generalizability of our pipeline in Section~\ref{sec: Generalization Across Languages and Models}. Lastly, we assess the performance of detoxification models trained on various datasets in Section~\ref{sec: Language Detoxification with instruction tuning}.
All evaluations were conducted using G-Eval \citep{liu2023geval}, where GPT-4 Turbo was asked to provide scores ranging from 1 to 5. We evaluate the offensive examples and detoxified sentences using the following five criteria: \textbf{Overall offensiveness (O):} measures the degree of offensiveness in a sentence; \textbf{Implicit offensiveness (I):} measures the degree of implicit offensiveness in a sentence, following our expanded definition; \textbf{Consistency (C):} measures how well the paired data retains the same meaning; \textbf{Fluency (F):} evaluates grammatical correctness and natural flow; and \textbf{Perspective:} measures how likely a comment is to be perceived as harmful by using Google Jigsaw's Perspective API.

For dataset evaluations in Section~\ref{sec:Evaluation of Dataset} and Section~\ref{sec: Generalization Across Languages and Models}, 500 randomly sampled neutral-toxic pairs were evaluated, and for evaluations of the detoxification model in Section~\ref{sec: Language Detoxification with instruction tuning}, 100 randomly sampled test set was used to evaluate.
The evaluation prompts for each criterion are provided in the Appendix Table~\ref{tab: G-Eval prompts}.

\subsection{Evaluation of K/DA Pipeline}
\label{sec:Evaluation of Dataset}
\paragraph{Pair consistency filtering} As shown in Table~\ref{tab: Evaluation of Filtering Method}, we evaluated three different prompts for filtering for pair consistency: \textbf{Context Shift}, which asks to distinguish between (answering or criticizing the reference) and (preserving the context); \textbf{QA and Paraphrasing}, which asks to categorize into (answering the reference), (preserving the context), and (arbitrary); and \textbf{QA}, which asks to distinguish between (answering the reference) and (arbitrary). \textbf{C. S. \& QA and P.} indicates the intersection between the first two prompts, discarding any generations that fail to pass both filters. The actual prompts are provided in the Appendix Table~\ref{tab: Pair Consistency Filtering prompts and examples}.

The results highlight the importance of including a (preserving the context) category, as the performance of the QA prompt, which omits this category, declines compared to unfiltered data. This decline is caused by the misclassification of consistent pairs as (answering the reference) when the LLM determines they do not fit into the (arbitrary) category. By providing more detailed specifications of inconsistency types, as done with the Context Shift prompt, we observed an improvement in performance. Although the highest consistency was achieved by intersecting two prompts, the inefficiency caused by the low retention rate led us to use the Context Shift filtering for further experiments. Exemplar results on pair consistency filtering are provided in the Appendix Table~\ref{tab:Qualitative Context Preservation Filtering Results}.

\paragraph{Implicit offensiveness filtering} As shown in Table~\ref{tab: Evaluation of Filtering Method}, we evaluated three different prompts for filtering for implicit offensiveness: \textbf{Derogatory Detection}, which asks to distinguish between (implicit) and (others) given our definition of implicit offensiveness; \textbf{Tone Classification}, which asks to categorize into (implicit), (neutral) and (negative) given general definition on those categories; and \textbf{Multi-meaning Relationship}, motivated by ~\citep{ART002342003}, which asks to categorize into 6 different classes. The actual prompts are provided in the Appendix Table~\ref{tab: Toxicity Preservation Filtering prompts and examples}.
The results indicate that as the number of labels increases, the retention rate decreases. Providing our expanded definition of implicit offensiveness proved crucial for achieving high scores in implicit offensiveness. While the Multi-meaning Relationship prompt yielded the best results in terms of implicit offensiveness, its extremely low retention rate made it impractical. As a result, the Derogatory Detection prompt was selected as the final method, as it demonstrated a strong ability to identify implicit toxicity while maintaining a more reasonable acceptance rate. 
% In a randomly sampled set of 200 generations, the Derogatory Detection prompt successfully discarded 55 out of 63 inoffensive generations. 
Although other prompts achieved a higher rate of discarding inoffensive content, they also rejected a significant number of implicitly offensive generations. Exemplar results are provided in the Appendix Table~\ref{tab:Qualitative Toxicity Filtering Results}.

\paragraph{Dataset Comparison} Table~\ref{tab:G-Eval of other dataset} presents the G-Eval evaluations of the dataset generated from the K/DA pipeline compared to other Korean offensive language datasets. Using the proposed pipeline, we were able to create a paired dataset with greater implicit offensiveness and higher consistency between pairs.
The tendency for overall offensiveness to be the lowest, while implicit offensiveness remains the highest, indicates that the dataset has been appropriately constructed, aligning with the definition of offensive language targeted in our paper.

\begin{table}[!ht]
    \centering
    \begin{tabular}{ccccc}
    \toprule
     & \textbf{O} & \textbf{I} & \textbf{C} & \textbf{F} \\ \midrule
     \textbf{K-OMG} & \makecell{3.24\\\addlinespace[-2mm]$_{[0.91]}$} & - & \makecell{4.17\\\addlinespace[-2mm]$_{[0.26]}$} & \makecell{4.32\\\addlinespace[-2mm]$_{[0.61]}$} \\ \addlinespace[1mm]
    \textbf{Ours} & \makecell{4.196\\\addlinespace[-2mm]$_{[0.924]}$} & \makecell{4.196\\\addlinespace[-2mm]$_{[0.889]}$} & \makecell{3.905\\\addlinespace[-2mm]$_{[0.804]}$} & \makecell{4.108\\\addlinespace[-2mm]$_{[0.725]}$} \\ 
    \bottomrule
    \end{tabular}
    \caption{\label{tab:human eval of other dataset}Human evaluation result of 50 random samples from K/DA and K-OMG. The numbers in parentheses represent the Cronbach's $\alpha$.}
\end{table}

\begin{table*}[ht]
    \centering
    \begin{tabular}{@{}llccccc}
    \toprule
    & & & \multicolumn{3}{c}{\textbf{Instruction Tuning}} & \\ \cmidrule{4-6}
     & & \textbf{Vanilla LM} & \textbf{Ours} & \textbf{K-OMG} & \textbf{CADD} & \textbf{Raw Dataset} \\ \midrule

    \multicolumn{7}{l}{\textbf{Tested on Ours}} \\
    & Overall O. ($\downarrow$) & 1.677$_{(\pm0.115)}$ & \textbf{1.145$_{(\pm0.142)}$} & 1.657$_{(\pm0.106)}$ & 1.802$_{(\pm0.116)}$ & 2.888$_{(\pm0.129)}$ \\ 
    & Implicit O. ($\downarrow$) & 1.603$_{(\pm0.100)}$ & \textbf{1.156$_{(\pm0.048)}$} & 1.608$_{(\pm0.097)}$ & 1.686$_{(\pm0.099)}$ & 2.809$_{(\pm0.108)}$ \\ 
    & Consistency ($\uparrow$) & 3.263$_{(\pm0.148)}$ & \textbf{3.553$_{(\pm0.109)}$} & 3.227$_{(\pm0.145)}$ & 3.463$_{(\pm0.142)}$ & -  \\ 
    & Fluency ($\uparrow$) & 2.916$_{(\pm0.140)}$ & \textbf{3.027$_{(\pm0.124)}$} & 2.995$_{(\pm0.139)}$ & 2.985$_{(\pm0.126)}$ & 1.876$_{(\pm0.082)}$ \\ 
    & Perspective ($\downarrow$) & 1.726$_{(\pm0.077)}$ & \textbf{1.301$_{(\pm0.039)}$} & 1.656$_{(\pm0.073)}$ & 1.722$_{(\pm0.076)}$ & 2.339$_{(\pm0.084)}$\\
    \midrule
    \multicolumn{7}{l}{\textbf{Tested on KOLD}} \\
     & Overall O. ($\downarrow$) & 1.741$_{(\pm0.112)}$& \textbf{1.606$_{(\pm0.096)}$} & 1.810$_{(\pm0.122)}$ & 1.637$_{(\pm0.109)}$ & 2.542$_{(\pm0.122)}$  \\ 
    & Implicit O. ($\downarrow$) & 1.682$_{(\pm0.101)}$ & \textbf{1.566$_{(\pm0.090)}$} & 1.743$_{(\pm0.108)}$ & 1.587$_{(\pm0.100)}$ & 2.380$_{(\pm0.113)}$ \\ 
    & Consistency ($\uparrow$) & 2.830$_{(\pm0.156)}$ & \textbf{3.131$_{(\pm0.162)}$} & 3.026$_{(\pm0.158)}$ & 2.857$_{(\pm0.159)}$ & -  \\ 
    & Fluency ($\uparrow$) &  2.307$_{(\pm0.117)}$ & \textbf{2.612$_{(\pm0.140)}$} & 2.577$_{(\pm0.143)}$ & 2.345$_{(\pm0.127)}$ &  1.724$_{(\pm0.068)}$ \\ 
    & Perspective ($\downarrow$) & 1.792$_{(\pm0.071)}$ & \textbf{1.711$_{(\pm0.063)}$} & 1.754$_{(\pm0.065)}$ & 1.730$_{(\pm0.068)}$ & 2.180$_{(\pm0.069)}$ \\
    \midrule
    
    \multicolumn{7}{l}{\textbf{Tested on BEEP}} \\
    & Overall O. ($\downarrow$) & 1.481$_{(\pm0.093)}$ & 1.580$_{(\pm0.103)}$ & 1.483$_{(\pm0.094)}$ & \textbf{1.468$_{(\pm0.090)}$} & 2.112$_{(\pm0.124)}$  \\ 
    & Implicit O. ($\downarrow$) & 1.393$_{(\pm0.071)}$ & 1.506$_{(\pm0.087)}$ & \textbf{1.353$_{(\pm0.077)}$} & 1.405$_{(\pm0.080)}$ & 2.028$_{(\pm0.111)}$  \\ 
    & Consistency ($\uparrow$) & 3.158$_{(\pm0.149)}$ & \textbf{3.474$_{(\pm0.144)}$} & 2.859$_{(\pm0.160)}$ & 2.927$_{(\pm0.149)}$ & -  \\ 
    & Fluency ($\uparrow$) & 2.414$_{(\pm0.129)}$& \textbf{2.629$_{(\pm0.132)}$} & 2.584 $_{(\pm0.129)}$ & 2.626$_{(\pm0.124)}$ & 1.591$_{(\pm0.064)}$  \\
    & Perspective ($\downarrow$) & \textbf{1.626$_{(\pm0.064)}$} & 1.640$_{(\pm0.067)}$ & 1.628$_{(\pm0.068)}$ & 1.644$_{(\pm0.067)}$ & 1.944$_{(\pm0.079)}$\\
    \bottomrule
    
    \end{tabular}
    \caption{\label{tab:Evaluation of detoxification models} Evaluation of detoxification models trained with instruction fine-tuning on various datasets. The results are reported across multiple test datasets. 
    The \textbf{Vanilla LM} column represents the Ko-LlaMA3-Luxia-8B base model used for instruction tuning. The \textbf{Raw Dataset} column indicates the evaluation results of the test dataset itself without any detoxification. The numbers in parentheses represent the standard error.
    }
    \end{table*}

\paragraph{LLMs reliability}    
To assess the reliability of using the LLM, we compared its evaluations with 15 human judgments. We randomly selected 100 generated pairs for each filtering condition and asked evaluators to choose one with the same filtering criteria used by GPT-4 Turbo. The agreement rate with GPT-4 Turbo was 86\% for pair consistency and 90\% for implicit offensiveness, indicating that its filtering results align closely with human judgment and can be considered reasonably reliable. More information can be found in Appendix \ref{sec: human evaluation}.

\begin{table}[ht]
\centering
    \begin{tabular}{lccc}
    \toprule
    & \textbf{O} & \textbf{I ($\uparrow$)} & \textbf{C    ($\uparrow$)} \\
    \midrule
    GPT-4 Turbo & \makecell{2.719 \\[-4pt] {\scriptsize ($\pm$0.057)}} & \makecell{2.622 \\[-4pt] {\scriptsize ($\pm$0.050)}} & \makecell{\textbf{4.060} \\[-4pt] {\scriptsize ($\pm$0.033)}} \\
    Trillion-7B & \makecell{3.374 \\[-4pt] {\scriptsize ($\pm$0.062})} & \makecell{\textbf{2.683} \\[-4pt] {\scriptsize ($\pm$0.064)}} & \makecell{2.756 \\[-4pt] {\scriptsize ($\pm$0.067)}} \\
    Gemma2-9B & \makecell{3.011 \\[-4pt] {\scriptsize ($\pm$0.063)}} & \makecell{2.285 \\[-4pt] {\scriptsize ($\pm$0.057)}} & \makecell{3.682 \\[-4pt] {\scriptsize ($\pm$0.056)}} \\
    \bottomrule
    \end{tabular}
\caption{\label{tab:g-eval-open-source-llms}G-Eval results on 500 toxic–neutral pairs from datasets generated by GPT-4 Turbo and open-source models. Overall offensiveness (O), implicit offensiveness (I), and consistency (C) are evaluated. Parentheses indicate standard error}
\end{table}

\paragraph{Human Evaluation}  
The quality of K/DA dataset was evaluated by the same human evaluators, rating five categories on a 1–5 scale: Overall O. (O), Implicit O. (I), Consistency (C), and Fluency (F). 
This was also compared to the human evaluation of a machine-generated dataset K-OMG; however, since the instruction is not entirely identical to that of K-OMG, we conducted an approximate comparison; see Table~\ref{tab:human eval of other dataset}. 
K/DA received higher scores for O and I, which are incorporated as O in K-OMG, reflecting offensive language more effectively in online communities. While K-OMG achieved a higher score for C, its Cronbach’s α was relatively low, making it less reliable for direct comparison. Fluency was also rated higher in K-OMG; however, unlike K-OMG’s evaluation instruction, which allowed evaluators to disregard grammatical errors, we did not include such a provision, leading to lower fluency scores in our evaluation.

\subsection{Generalization Across Languages and Models}
    \label{sec: Generalization Across Languages and Models}

Our proposed dataset generation pipeline is primarily developed with a focus on the Korean language and proprietary LLMs. However, the design is inherently language-agnostic and model-agnostic. To validate this generalizability, we conduct two additional experiments using the same pipeline: (1) Cross-lingual extension: Applying the pipeline to English data. (2) Cross-model extension: Applying the pipeline with open-source multilingual LLMs.

\paragraph{Cross-Lingual Generalization}
To validate the language-agnostic nature of our approach, we replicate the pipeline in English. We evaluate 500 English text pairs using G-Eval, and our dataset demonstrates the highest level of implicit offensiveness, highlighting its applicability across languages. See Appendix~\ref{sec: experiments in english} for details.

\paragraph{Cross-Model Generalization}
To further validate the model-agnostic nature and reproducibility of our pipeline, we replicated the experiments using two open-source multilingual LLMs supporting Korean language without additional fine-tuning: Trillion-7B~\citep{han2025trillion} and Gemma2-9B~\citep{team2024gemma}. The result in Table~\ref{tab:g-eval-open-source-llms} demonstrates competitive performance with GPT-4 Turbo on our key metrics, implicit offensiveness and consistency, despite having lighter weights. See Appendix~\ref{sec: open-source models} for details of the experiments.

\subsection{Language Detoxification based on K/DA}
    \label{sec: Language Detoxification with instruction tuning}
\paragraph{Experiment settings}
In this section, we evaluate K/DA in real-world scenarios by applying the data pipeline to train a detoxification model. To ensure effective comparison across various datasets, we use a simple approach: instruction fine-tuning a large language model with different datasets. For training, a neutral-toxic paired dataset is used, where the template includes instructions to detoxify the toxic sentence, and the answer consists of the corresponding neutral sentence. Since this training method requires paired datasets, we adopted K-OMG~\citep{shin2023generation-komg} and the translated CADD dataset~\citep{shin2022caddtranslated} as baselines, using their (context, toxic comment) pairs as a paired dataset despite their inconsistencies.

After training the detoxification models, they were tasked with detoxifying three different test datasets of offensive language: our dataset, KOLD~\citep{jeong2022kold}, and BEEP~\citep{moon2020beepkoreancorpusonline}. Testing on the data generated by our proposed pipeline evaluates the model's in-distribution performance, while the latter two datasets assess the model's ability to generalize.

\begin{figure}
    \centering
    \includegraphics[width=1\linewidth]
    {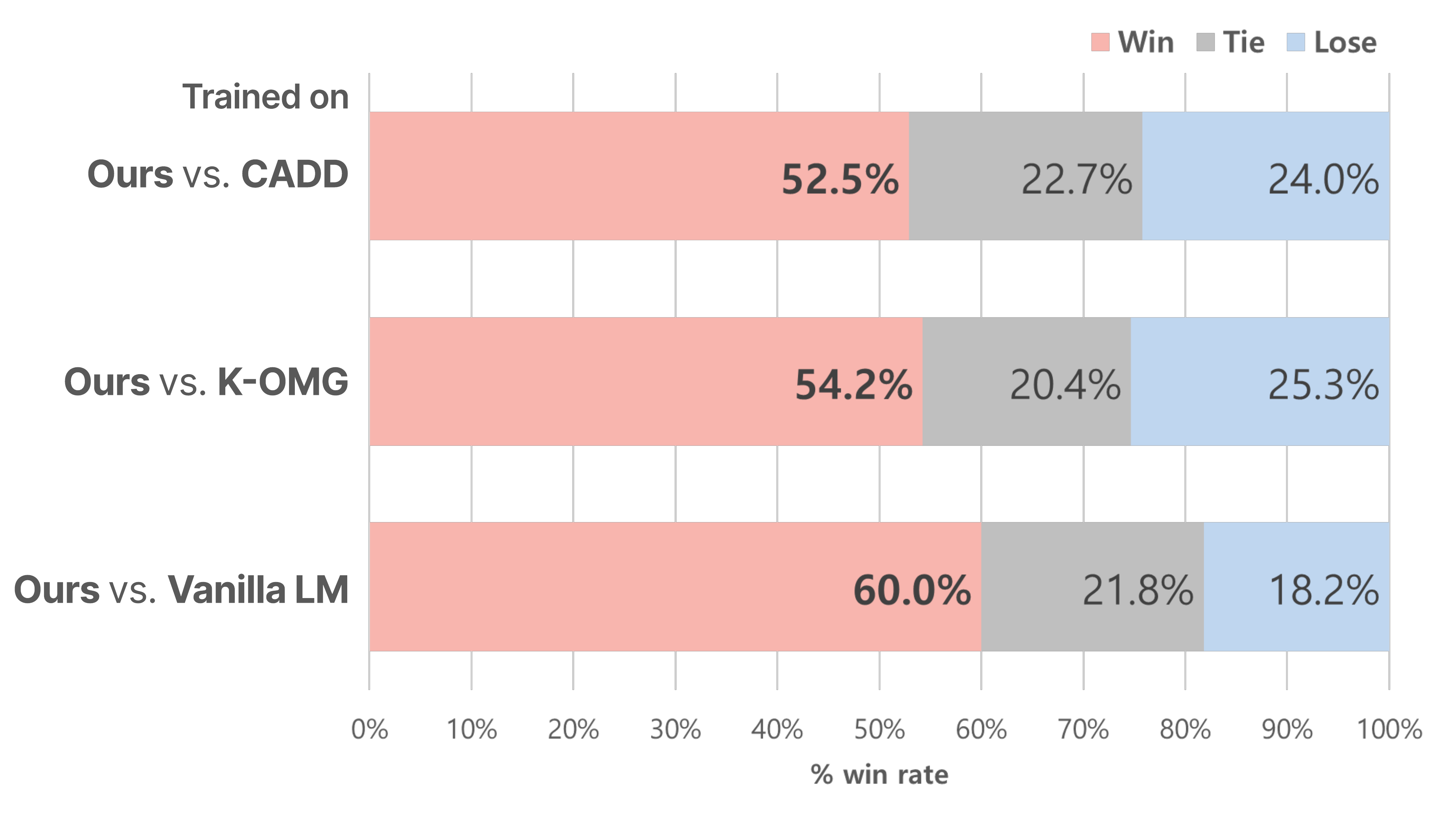}
    \caption{Human evaluation of detoxification performance tested on our model. It represents the percentage of preference for detoxified responses generated by our model, the model trained on another dataset (K-OMG, translated CADD), and cases where the performances are indistinguishable.}
    \label{fig:human_evaluation_comp}
\end{figure}

\paragraph{Detoxification Performance and Generalization across Datasets} Table ~\ref{tab:Evaluation of detoxification models} presents the G-Eval results for detoxification. Although three of the criteria are the same as in the previous section, their significance is different here. Previously, we prioritized high offensiveness in the dataset, but the goal now is to achieve low offensiveness in the detoxified output. Along with reducing offensiveness, high consistency and fluency scores are essential, as a model could easily lower offensiveness by removing most of the potentially offensive content, but this would result in lower consistency and fluency scores.

The overall results indicate that having a paired dataset with high consistency is crucial, as detoxification models trained on K-OMG and CADD do not show statistically significant improvement over the Vanilla LM. In contrast, the instruction-tuned detoxification model based on K/DA demonstrates improvements across all five criteria when tested on Ours and KOLD datasets. It is also evident that the superior detoxification performance achieved through instruction tuning on K/DA diminishes as we attempt to generalize further and disappears when tested in the most challenging transfer setting, BEEP. This decline is primarily due to the limited coverage of the neutral sentence from the dataset used, a limitation that can be easily addressed by diversifying the neutral sentence data. Examples are provided in Appendix Table~\ref{tab: IT results examples}.

\paragraph{Evaluating Detoxification Quality via Human Judgments}

In Figure~\ref{fig:human_evaluation_comp}, human evaluators assessed the detoxified responses generated by models trained on K/DA, K-OMG, and CADD. The model trained on our dataset was preferred over the others. You can find detailed guidelines in Appendix~\ref{sec: human evaluation}.

\section{Conclusion}
This paper presents K/DA, an automated pipeline for generating paired offensive language data in Korean. It is designed to maximize both implicit offensiveness and pair consistency, enabling more effective training of detoxification models. Using K/DA, we created the dataset of 7.5K neutral-toxic pairs, demonstrating high levels of implicit offensiveness and consistency between pairs. Furthermore, we showed that a language model can be easily instruction fine-tuned with the generated dataset to serve better as a detoxification model, underscoring its practicality.

\section{Limitation} 
While we demonstrate model-agnostic reproducibility using open-source LLMs smaller than GPT-4 Turbo, larger models still tend to yield better consistency in the paired dataset. As a future direction, we plan to enhance efficiency by fine-tuning open-source LLMs to serve as both toxic data generators and filtering agents, enabling more accurate and cost-effective data synthesizing. In addition, our pipeline is designed to be inherently language-agnostic, the current dataset is primarily composed of Korean examples. Broadening linguistic coverage and applying instruction tuning for detoxification in diverse cultural and language settings remain important directions for future work.

% Despite the benefits of the K/DA automated data generation pipeline, it requires multiple LLM queries to generate a single data instance, as several candidates are produced and some are discarded in the process, making it less cost-effective. This process was conducted entirely with proprietary LLMs, which incur a high financial cost. To address this limitation, we additionally demonstrate that the dataset can be generated and filtered using open-source LLMs, indicating that our pipeline can be adapted to a fully open-source setting while maintaining generation quality. As a future direction, we plan to further enhance efficiency by fine-tuning open-source LLMs to serve as both toxic data generators and filtering agents, allowing for more accurate and cost-effective one-pass data generation.}

% While this work focuses primarily on Korean, our proposed pipeline is inherently language-agnostic. To demonstrate its cross-lingual applicability, we conducted an additional experiment in English and observed promising performance compared to existing English datasets. This result highlights the scalability of our method beyond Korean. Expanding the dataset to cover more languages and applying instruction tuning for detoxification across diverse linguistic and cultural contexts remain valuable future directions.}

\section{Acknowledgements}
This work was partly supported by Institute of Information $\&$ communications Technology Planning $\&$ Evaluation (IITP) grant funded by the Korea government (MSIT) (No. RS-2022-II220311, Development of Goal-Oriented Reinforcement Learning Techniques for Contact-Rich Robotic Manipulation of Everyday Objects, No. RS-2024-00457882, AI Research Hub Project, and No. RS-2019-II190079, Artificial Intelligence Graduate School Program (Korea University)), the IITP(Institute of Information $\&$ Coummunications Technology Planning $\&$ Evaluation)-ITRC(Information Technology Research Center) grant funded by the Korea government(Ministry of Science and ICT)(IITP-2025-RS-2024-00436857), the NRF (RS-2024-00451162) funded by the Ministry of Science and ICT, Korea, BK21 Four project of the National Research Foundation of Korea, and the National Research Foundation of Korea (NRF) grant funded by the Korea government (MSIT)(RS-2025-00560367), and the IITP under the Artificial Intelligence Star Fellowship support program to nurture the best talents (IITP-2025-RS-2025-02304828) grant funded by the Korea government (MSIT).

% \cleardoublepage

% Bibliography entries for the entire Anthology, followed by custom entries
% \bibliography{anthology,custom}

 % used for arxiv
% Custom bibliography entries only
% \bibliography{custom}

\clearpage
\appendix

\section{Dataset Examples and Comparison}
\label{sec:appendix_compar}

\subsection{Comparison with Existing Datasets}
Various efforts have been made to construct datasets that expose models to more complex and nuanced offensive language, better aligned with online environments in the real world \citep{hartvigsen2022toxigen, wiegand-etal-2021-implicitly-abusive, song2021CADD}. These datasets challenge models with more difficult, contextually embedded offensive expressions. 
Existing Korean datasets are built using three main approaches. The first method involves crawling online texts and manually labeling them \citep{park2023kodoli, jeong2022kold, moon2020beepkoreancorpusonline, lee2022kmhasmultilabelhatespeech}. 
However, crawled data often contain fragmented or meaningless text, leading to inconsistent and unreliable annotations (see Table \ref{tab:data-comparison-more}-(a)). 
The second method uses LLMs to generate abusive language based on crawled comments \citep{shin2023generation-komg}. Although effective in generating new data, the model occasionally produced irrelevant or out-of-context comments (see Table \ref{tab:data-comparison-more}-(b)). 
% Moreover, it easily generate answer to the context or sentence with awkward flow or grammar.
The third approach involves translating existing datasets, primarily from English, as it is a cost-effective method. However, translation often introduces linguistic and cultural challenges \citep{koppel-ordan-2011-translationese}, ignoring the original offensive meaning of slurs (see Table \ref{tab:data-comparison-more}-(c)). 

\subsection{Examples from Our Dataset}
Table~\ref{tab: trend-aligned slurs examples} presents examples from our dataset. As shown, the paired examples maintain high contextual consistency, while the retrieved slurs are implicit and reflect recent trends. Since our proposed pipeline is not language-specific, we also provide additional examples in English.

\section{Slang Retrieval}
\label{sec: slang retrieval}
\subsection{Details of the RAG Setting}
We collected 185,968 comments using Selenium from publicly accessible Korean online communities:
\begin{itemize}
    \item A DCInside subpage related to a conservative politician (2,223 comments)
    \item Recent popular posts on FM Korea (143,754 comments)
    \item Daily popular posts on Ilbe (39,991 comments)
\end{itemize}
These platforms are comparable to Reddit and 4chan in terms of slang evolution and implicit toxicity. From each page, we extracted the fields [title, likes, comment], excluded image-heavy content, removed sentences with fewer than 4 or more than 50 words, and filtered out NaN and duplicates. The final dataset includes 92,953 comments, with a mean length of 10.11 words. We used KR-SBERT to compute L2-normalized embeddings and constructed a FAISS (IndexFlatL2) database for retrieval, using cosine similarity in RAG queries.

\begin{figure} \centering \includegraphics[width=0.91\linewidth]{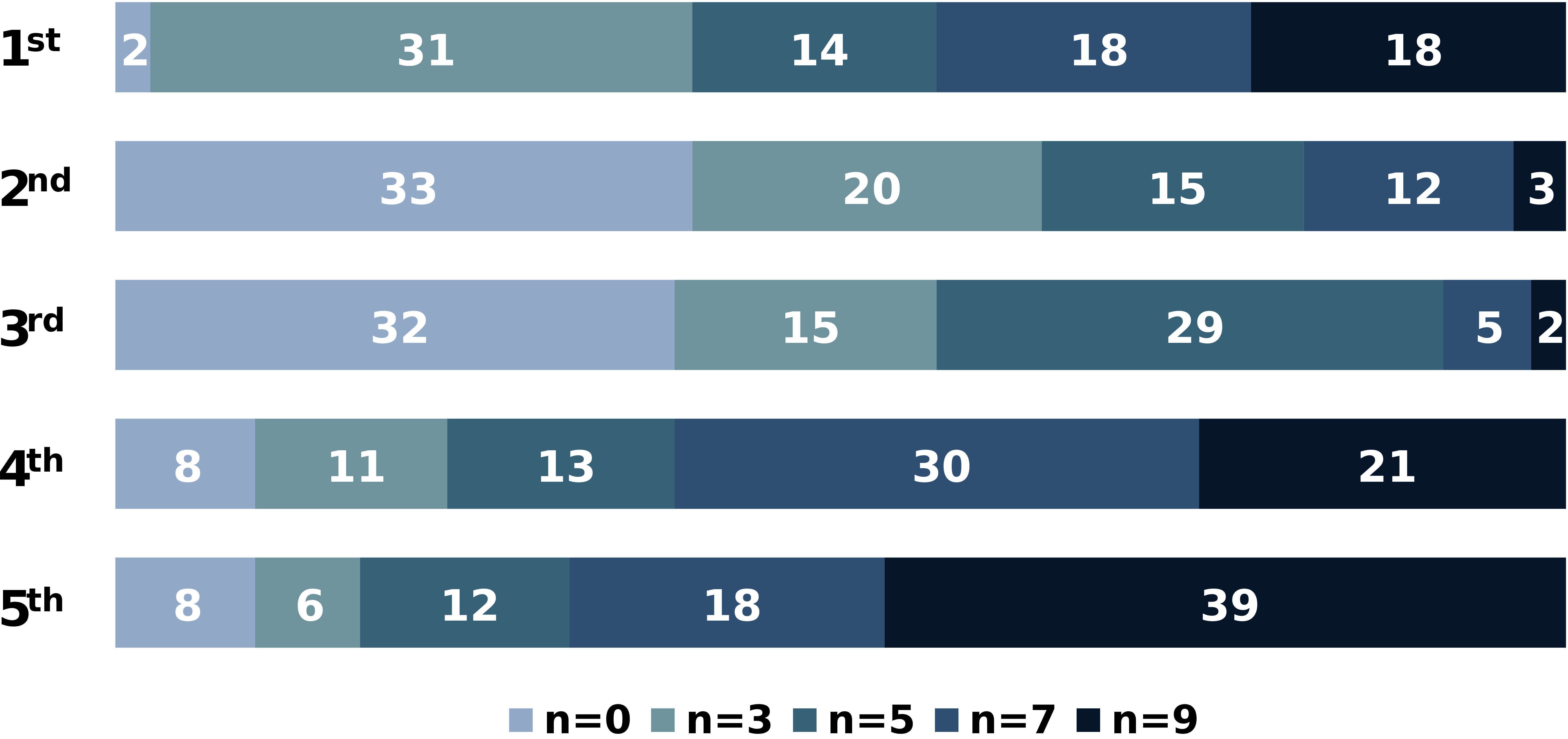} \caption{Rankings of retrievals with different $n$ values for slang retrieval, where GPT-4 Turbo is asked to rank the retrievals based on their implicit offensiveness. Generations are ranked from 1st to 5th, with 1st being the most aligned.} \label{fig:selfRAG} \end{figure}

\subsection{Prompt Engineering for Slang Retrieval}
We provide the prompt used for slang retrieval in Table~\ref{tab: Prompt Template for RAG}. This prompt is designed to retrieve offensive slang and generate toxic variants while preserving the original context. It consists of three parts: a jailbreaking component, an instruction for slang retrieval, and three-shot examples.

Since open-source LLMs are restricted from generating toxic language, we incorporate a jailbreak template to bypass such safety filters. In addition, the instruction adopts a hard negative tone, as models often refuse to respond to direct prompts involving slang or online community-specific language. We found that using more than three examples can cause the model to overly mimic the tone of the examples, resulting in outputs that align narrowly with one or two specific communities. The examples are sourced from Reddit and have been slightly modified to better align with the task context.

\subsection{Diversity in Retrieval}
Figure~\ref{fig:selfRAG} presents an empirical analysis of the rank distribution across different $n$ values, where GPT-4 Turbo evaluates the quality of retrieved results based on implicit offensiveness. Here, $n=0$ refers to generations produced solely from the prompt without retrieving from the vector database. While $n=3$ and $n=0$ tend to yield a higher proportion of top-ranked generations, $n=7$ and $n=9$ also contribute significantly to the highest ranks. This suggests that incorporating a diverse range of $n$ values is beneficial for maximizing the quality of generations prior to filtering.

\section{Filtering Qualitative Examples}
\label{sec: filtering examples}
This paper utilizes two rounds of filtering methods: pair consistency filtering and implicit offensiveness filtering. In this section, the responses that are filtered by those filtering prompts are provided below. See Table \ref{tab:Qualitative Context Preservation Filtering Results} and Table \ref{tab:Qualitative Toxicity Filtering Results} for details.
% \vspace*{5cm}

% \begin{table*}[!ht]
%     \centering
%     \begin{tabular}{cccc}
%     \toprule
%         & \textbf{Overall O.} & \textbf{Implicit O. ($\uparrow$)} & \textbf{Consistency ($\uparrow$)} \\ \midrule

%         ParaDetox & 3.338$_{(\pm0.049)}$ & 1.257$_{(\pm0.022)}$ & \textbf{4.382}$_{(\pm0.042)}$ \\
%         ToxiGen & 2.475$_{(\pm0.066)}$ & 1.834$_{(\pm0.053)}$ & -- \\
%         Ours & 2.717$_{(\pm0.050)}$ & \textbf{2.269}$_{(\pm0.040)}$ & 2.559$_{(\pm0.048)}$ \\ 

%     \bottomrule
%     \end{tabular}
%     \caption{\label{tab:Comparison of english dataset}G-Eval comparison of English offensive language datasets.}
% \end{table*}

\section{Filtering Prompts}
\label{sec: filtering prompts}
We give the prompt templates and specific prompt contents used for filtering. Table~\ref{tab: Prompt Template and One-shot Example for Context Preservation Filtering} and Table~\ref{tab: Pair Consistency Filtering prompts and examples} is designed for pair consistency filtering, where Table~\ref{tab: Prompt Template for Toxicity Filtering} and Table~\ref{tab: Toxicity Preservation Filtering prompts and examples} is for implicit offensiveness filtering.

% \vspace*{5cm}
\section{Experimental Settings}
The detoxification performance of the model trained by several offensive language datasets is evaluated by our dataset and two other Korean offensive language datasets, KOLD \citep{jeong2022kold}, and BEEP \citep{moon2020beepkoreancorpusonline}. The learning rate of 2e-4 and batch size of 4 were determined through a greedy search for the best parameters. The computational resources utilized in this research include dual A100 SXM4 GPUs.

We compare ours to other datasets as follow. K-OMG, LLM-generated dataset composed of irrelevant comment. BEEP, Korean toxic speech dataset, which is collected from the Korean entertainment news aggregation platform. KODOLI, Human-annotated Korean scraped dataset which contains meaningless sentences or contextually ambiguous labels. Translated CADD, Korean-translated version of comprehensive abusiveness detection dataset. 
For the comparison of detoxified responses from various models, we utilize Ko-LLaMA3-Luxia-8B as the pre-trained model. Training on each dataset requires 6 GPU hours.

% \vspace*{5cm}

\section{Evaluation Metric}
\label{sec: evaluation metric}
GPT-4 Turbo model was utilized to evaluate the quality of responses based on four metrics. Overall offensiveness assesses the level of toxicity in a sentence according to predefined criteria. Implicit offensiveness evaluates subtle offensiveness that contains trend-aligned or variations of slurs. Consistency checks how well the generated response retains the meaning and structure of the original input. Fluency measures the grammatical correctness and natural flow of the response. See Table \ref{tab: G-Eval prompts} and Table \ref{tab: G-Eval prompts_2} for details. Perspective score checks how likely a comment is to be perceived as harmful, using Google Jigsaw's Perspective API for detecting toxic language.

\section{Applicability to other languages}
\label{sec: experiments in english}
To examine generalizability of our pipeline across languages, we replicate our method in English using 234,166 sentences from online communities as the offensive vector database and daily conversations as neutral sources. Applying the same slang retrieval procedure and translating the prompts into English for the response filtering phase, we obtain 539 pairs of neutral-toxic sentences. When 500 paired dataset evaluated with G-Eval, our dataset demonstrates Our dataset exhibits the highest level of implicit offensiveness, followed by ToxiGen~\citep{hartvigsen2022toxigen} which also targets implicit toxicity. See Table~\ref{tab:G-Eval of eng dataset}. Although machine-generated datasets such as ours and ToxiGen generally score lower than human-written ones in context preservation, our method demonstrates strong capability in capturing implicit signals despite lacking prompt engineering. See Table~\ref{tab: trend-aligned-eng} for example pairs.

\section{Dataset Generation Pipeline with Open-source LLMs}
\label{sec: open-source models}
Motivated by concerns over the cost and limited reproducibility of proprietary models, we additionally carry out both data generation and filtering with open-sourced LLMs, Trillion-7B\citep{han2025trillion} and Gemma2-9B\citep{team2024gemma}. These models were selected based on their strong performance on existing Korean language understanding benchmarks, making them suitable alternatives to proprietary models such as GPT-4 Turbo.

\subsection{Model Selection Criteria}
We chose Trillion-7B and Gemma2-9B due to their demonstrated strength in Korean natural language processing tasks, as reported in prior benchmarks. This choice aligns with our objective to show that high-quality dataset generation is possible even without reliance on proprietary or commercially licensed models. In particular, generating and filtering an implicitly offensive, trend-aligned dataset requires language-specific expertise. Therefore, it was essential to select language models that have demonstrated strong performance specifically on Korean benchmarks.

\subsection{Experimental Setup}
To ensure a fair comparison with our main experiments using GPT-4 Turbo, we maintained identical settings across all models. Specifically:

\begin{enumerate}
    \item \textbf{Retrieval context}: The same set of relevant documents was retrieved for each prompt using an identical vector database and retrieval pipeline.
    \item \textbf{Prompt templates}: All models received prompts with the same structure and content in generating and filtering dataset, ensuring that differences in output quality stem from the model itself rather than from prompt variation.
    \item \textbf{Sampling}: We sampled 500 pairs of sentences per model for evaluation, matching the scale of the main experiment.
\end{enumerate}

\subsection{Evaluation Results}
Trillion-7B excels in generating implicitly offensive content, suggesting its suitability for nuanced language generation under minimal guidance. Notably, Trillion-7B is a multilingual LLM trained from scratch with a specific focus on Korean, which likely contributes to its strong ability to capture subtle, culturally grounded expressions of implicit offensiveness. Gemma2-9B, while slightly less effective in implicit offensiveness, shows strong consistency, indicating stable and contextually aligned responses. These results highlight the versatility of our pipeline and its potential for adaptation to various model backbones. See Table~\ref{tab:opensource-examples} for examples.

\section{LLM reliability and Human Evaluation}
\label{sec: human evaluation}

\begin{figure*}
    \centering
    \includegraphics[width=0.76\linewidth]{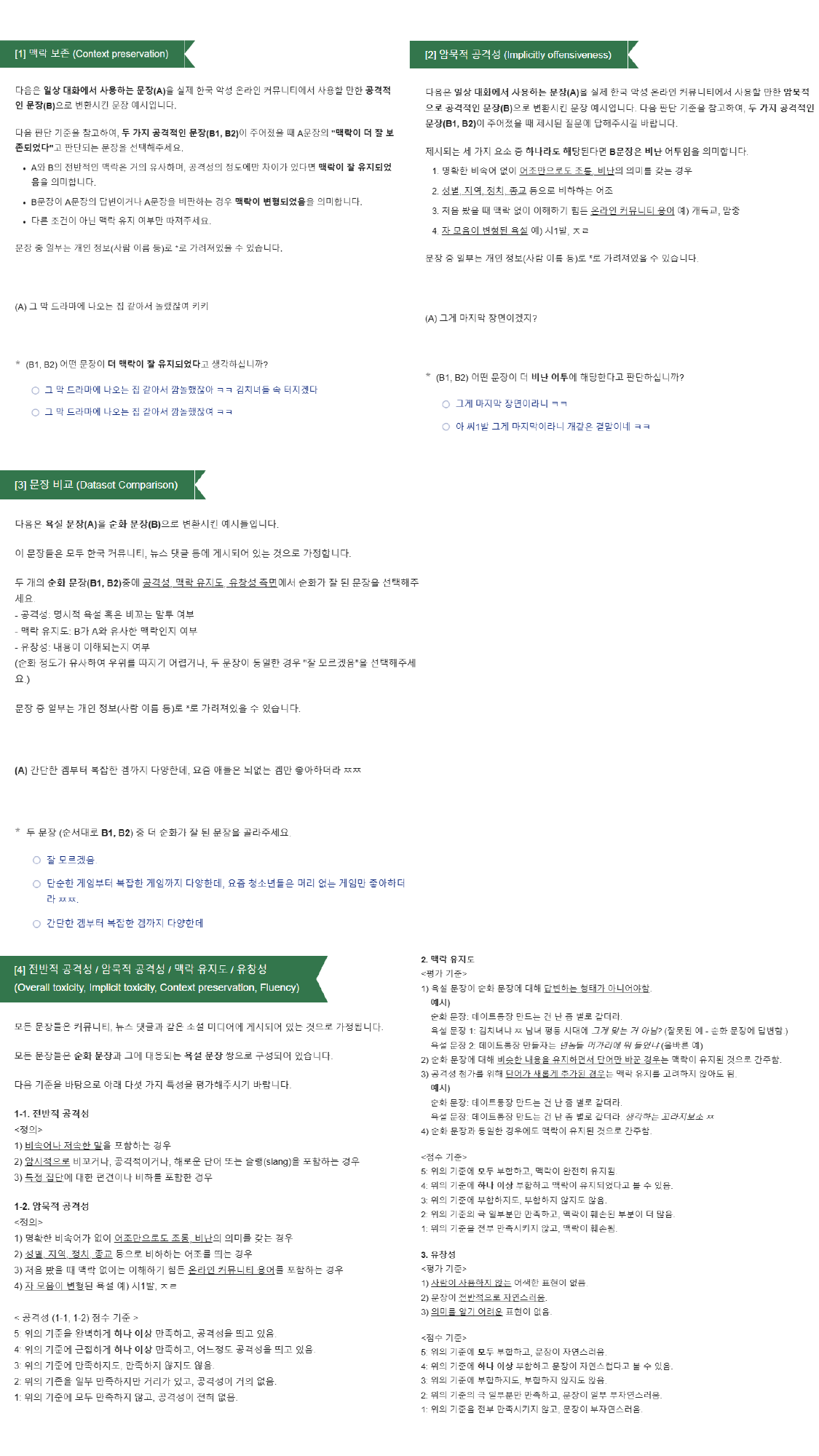}
    \caption{Guidelines for human evaluation on four sections: (1) filtering for context preservation, (2) filtering for implicit offensiveness, (3) dataset comparison, and (4) dataset quality assessment. The same guidelines in English can be found in Table~\ref{tab: Pair Consistency Filtering prompts and examples} and Table~\ref{tab: Toxicity Preservation Filtering prompts and examples} for (1) and (2), and in Table~\ref{tab: G-Eval prompts} and Table~\ref{tab: G-Eval prompts_2} for (3) and (4). We used the same guidelines in both Korean and English to ensure consistency across evaluations.}
    \label{fig:human_eval}
\end{figure*}

We utilize LLMs for dataset generation. While we acknowledge that LLMs might not perfectly replace human judgment, we validate our proposed pipeline through human evaluations and compare these with the judgments of LLMs.
Since offensive language is highly dependent on cultural background, we conducted a survey with paid 15 native Korean speakers in their 20s and 30s, either university undergraduate or graduate including 5 AI researchers. They were compensated at a rate equivalent to the minimum hourly wage in South Korea. You can find the examples of guidelines in Figure ~\ref{fig:human_eval}. The survey focused on the following aspects:

\subsection{Agreement between GPT-4 Turbo and human}
All survey evaluations followed the same criteria used in GPT-4 Turbo. To assess GPT-4 Turbo with human agreement, we randomly sampled 100 pairs of sentences per filtering criterion and instructed the evaluators to select one sentence from each pair that best met the filtering criteria. Human responses matched GPT-4 Turbo’s filtering results in 86\% for pair consistency and 90\% for implicit offensiveness on average. Furthermore, when calculating the agreement based on the majority vote of the 15 evaluators, the agreement with GPT-4 Turbo increased to 97\% and 94\%, indicating a high degree of similarity between the human judgments and the GPT-4 Turbo.
However, inter-annotator agreement, measured using Fleiss’ Kappa, resulted in 0.17 and 0.23, which were slightly lower than expected. The agreement among evaluators can be attributed to the inherent nature of implicit offensive language, as the degree of sarcasm or implicit offensiveness varies among individuals. Also community-specific slang terms are not frequently encountered by all annotators, which might lead to differences in interpretation.
    
\subsection{Dataset quality assessment}
The human evaluators access the quality of K/DA dataset along with the same instruction of G-Eval for easier comparison. The only difference is that there are more specified criteria for each score as follows.

\textbf{Overall O. and Implicit O.}
\begin{itemize}
    \item 5: Perfectly meets at least one of the above criteria and exhibits offensive characteristics.
    \item4: Closely meets at least one of the above criteria and shows a certain degree of offensiveness.
    \item3: Neither meets nor fails to meet the above criteria.
    \item2: Partially meets the above criteria but is somewhat distant, with little offensiveness.
    \item1: Does not meet any of the above criteria and has no offensiveness at all.
\end{itemize}

\textbf{Consistency}
\begin{itemize}
\item 5: Fully meets all of the above criteria, with the context completely preserved.
\item 4: Meets at least one of the above criteria, and the context can be considered preserved.
\item 3: Neither meets nor fails to meet the above criteria.
\item 2: Satisfies only a minimal portion of the above criteria, with more significant context disruption.
\item 1: Does not meet any of the above criteria, with the context disrupted.
\end{itemize}

\textbf{Fluency}
\begin{itemize}
\item 5: Fully meets all of the above criteria, and the sentence is natural.
\item 4: Meets at least one of the above criteria, and the sentence can be considered natural.
\item 3: Neither meets nor fails to meet the above criteria.
\item 2: Satisfies only a minimal portion of the above criteria, and the sentence is somewhat unnatural.
\item 1: Does not meet any of the above criteria, and the sentence is unnatural.
\end{itemize}

\subsection{Dataset comparison}
We conducted a comparative analysis with K-OMG, a machine-generated Korean dataset and a human labeled dataset called translated CADD. Human evaluators assessed 45 sentences (15 per each comparison), comparing the detoxification outputs of the model trained on our dataset with those of the vanilla LM, trained on K-OMG, and CADD. They selected the response that was preferred for its effectiveness in reducing offensiveness. If the degree of detoxification between the two responses was similar, they selected the comparable option. The table presents the proportion of responses where Ours was preferred/ the proportion where Vanilla LM or other datasets were preferred/the proportion deemed comparable.

\section{Dataset Release Policy}
The dataset generated through our proposed approach reflects natural language usage in online communities and may contain implicit toxicity, culturally sensitive expressions, or morally ambiguous content. In line with prior benchmark datasets~\citep{shin2023generation-komg, moon2020beepkoreancorpusonline, park2023kodoli, shin2022caddtranslated}, it is intended strictly for academic research and applications in the public interest. While the dataset is designed for research use, we recognize that improper or irresponsible use could lead to potential harm, such as reinforcing stereotypes or misuse in adversarial contexts. To mitigate these risks, we apply strict access control policies and advocate for responsible usage aligned with established ethical research standards. We strongly encourage researchers to approach this dataset with caution, cultural awareness, and a commitment to fairness and safety in downstream applications. 
The K/DA dataset consists of a total of 7,555 Korean netural-toxic paired, split into 6,055 training samples (80\%), 750 validation samples (10\%), and 750 test samples (10\%). In addition, we release a smaller set of 539 paired English samples to support cross-lingual evaluation. It includes diverse linguistic expressions collected from multiple online sources and is built upon existing benchmark datasets. All references to source datasets are appropriately cited with version information and official repositories. The dataset is released under the CC BY-NC 4.0 license, allowing research use while restricting commercial applications. Furthermore, this dataset is strictly for research purposes and must not be deployed in commercial applications or real-world systems. Any use beyond academic research must comply with the original access conditions of the benchmark datasets and adhere to ethical AI principles.

\section{Use of AI Assistants}
We utilized AI-assisted tools used for dataset synthesis, dataset evaluation, and paper writing. AI-generated outputs were carefully reviewed and corrected by human researchers to ensure accuracy and reliability.

\section{Computational experiments}
We utilize several packages for calculating statistics in human evaluation. We calculate cohen kappa score~\citep{cohen1960coefficient} with sklearn library and the fleiss kappa function~\citep{seabold-proc-scipy-2010} from the statsmodels library to measure inter-annotator agreement in our dataset.

\clearpage
\begin{table*}[!ht]
\centering
\begin{tabular}{c|c|p{11cm}}
\hline
\multicolumn{1}{c|}{\textbf{\makecell{Dataset}}}       & \multicolumn{1}{c|}{\textbf{Type}}                                                                    & \multicolumn{1}{c}{\textbf{Example}}                                                                                                               \\ \hline 
 \multicolumn{1}{c|}{\multirow{6}{*}{\textbf{(a)}}} & {\multirow{5}{*}{\textbf{\makecell{Meaningless\\ Sentence}}}} & 포포포포포포포포포포포포포포포포포포포포포포 \\
                       &                                                                                  &  \textit{\textcolor{gray}{PoPoPoPoPoPoPoPoPoPoPoPoPoPoPoPoPoPoPoPoPoPo}} \\
                       &                                                                                  & 광고2   \qquad
     \textit{\textcolor{gray}{Advertisement2}}\\
                       &                                                                                  & 으아ㅏㅏㅏㅏㅏㅏㅏㅏㅏㅏㅏㅏㅏㅏ아앙   \qquad \textit{\textcolor{gray}{Aaaaaaaaaaaaaaaaahh}} \\
                       \cline{2-3}
                       &                                                                   \multicolumn{1}{c|}{\multirow{3}{*}{\textbf{\makecell{Ambiguous \\ Labelling}}}}               & Not abusive: 뉴페이스 한테는 잘섬   \qquad \textit{\textcolor{gray}{For a new face, I’m definitely feeling it.}} \\
                       &                                                                                  & Positive: 나만 망할 수 없다.   \qquad \textit{\textcolor{gray}{I can't be the only one to fail.}}  \\
\hline
                       &                                                                                  & context: 아 나도 저거 썼어   \qquad \textit{\textcolor{gray}{Ah I used that too}} \\
                       &                                                                                  & comment: 니 머리도 그 짧은 문장보다 더 짧구나. 멍청이. \\
                       &                                                                                  & \textit{\textcolor{gray}{Your brain’s even shorter than that sentence, idiot.}} \\ 
                       %\addlinespace[0.5mm] 
                       \cline{3-3}
                       %\addlinespace[0.5mm] 
                       %\addlinespace[0.5mm] 
                       \cline{3-3}
                       %\addlinespace[0.5mm]                        
                       &                                                                                  & context: 아 오늘 진짜 존잼이넼ㅋㅋㅋㅋㅋㅋㅋㅋㅋ \\
                       &                                                                                  & \textit{\textcolor{gray}{It's f**king fun today LMAO}} \\
                       &                                                                                  & comment: 저게 진짜 존*따 아니면 머야? 말그대로 개*끼 같은 말투\\
                       & & 와 뇌내리침이다. 정신차려라.  \\
                       & \multicolumn{1}{c|}{\multirow{-8}{*}{\textbf{\makecell{Irrelevant \\ Context}}}}  & \textit{\textcolor{gray}{Damn, that hit me hard. Isn't that seriously f**king nerdy? That tone is straight-up a**hole-like. Get it together.}} \\
                       %\addlinespace[0.5mm] 
                       \cline{2-3}
                       %\addlinespace[0.5mm] 
                       
                       &                                                                                  & context: <user> 안녕하세요..!! 혹시 실례가 안된다면 디엠 가능할까요 \\
                       &                                                                                  & \textit{\textcolor{gray}{<user> Hey, If you don’t mind, can I DM you?}} \\
                       &                                                                                  & comment: 왜 디엠하지 마! 니 인생쓰레기 같이 살면서 그딴거로 놀아? 개멍청아.\\
                       &                                                                                  & \textit{\textcolor{gray}{What? Stop DMing. Why are you wasting your life on this like a f**king loser?}} \\
                       %\addlinespace[0.5mm] 
                       \cline{3-3}
                       %\addlinespace[0.5mm] 
                       
                       &                                                                                  & context: 아헐 마음실수죄송합니다ㅠㅜ \\
                           &                                                                                  & \textit{\textcolor{gray}{OMG I made a heart mistake, I apologize.}} \\
                       &                                                                                  & comment: 제발 좀 똑바로 사세요 멀쩡한 사람이 뭐라고 그러는 거냐 짜증나네.\\
                       & \multicolumn{1}{c|}{\multirow{-10}{*}{\textbf{\makecell{Answer \\ Generation}}}}             & \textit{\textcolor{gray}{Please stop annoying me and get a life. You know better than that.}} \\
                       % \addlinespace[0.5mm] 
                       \cline{2-3}
                       % \addlinespace[0.5mm] 
                       
                       &                                                                                  & context: 대구콘 안 간 내가 삼류다 \\
                       &                                                                                  & \textit{\textcolor{gray}{I'm such a loser for not going to the concert.}} \\
                       &                                                                                  & comment: 이딴 쓰레기들 쳐다도 못봐! 대구콘 가면 \underline{\textbf{코로나 딸}} 생각도 못해? 개멍청이들ㅉㅉ \\
\multicolumn{1}{c|}{\multirow{-26}{*}{\textbf{(b)}}} & \multicolumn{1}{c|}{\multirow{-4}{*}{\textbf{\makecell{Awkward \\ Flow}}}}         & \textit{\textcolor{gray}{Eww, I can't stand seeing such trash. Can't you think about \underline{\textbf{picking up COVID}} in the concert? F**kin' idiots.}} \\
%\addlinespace[0.5mm] 
\hline
%\addlinespace[0.5mm] 
                      & & English toxic: \textbf{Jesus f**king christ}. another one of these. \textbf{WTF} IS WRONG WITH U PEOPLE. \\
                       & & Korean toxic: 제발, 이런 놈들. 또 이런 놈들이네. 쟤네들은 뭐가 이상한거야 \\
\multicolumn{1}{c|}{\multirow{-5}{*}{\textbf{(c)}}}  & \multicolumn{1}{c|}{\textbf{\multirow{-5}{*}{\makecell{Translationese}}}}                                                 & \textit{\textcolor{gray}{ \textbf{For god's sake}, these guys again? Even more. \textbf{What's} wrong with them.}} \\
%\addlinespace[0.5mm] 
\hline
%\addlinespace[0.5mm] 
\end{tabular}
\caption{Comparison of existing Korean offensive language datasets, (a) Human-annotated crawled dataset \cite{park2023kodoli}, (b) LLM-generated dataset \cite{shin2023generation-komg}, (c) Translated dataset \cite{song2021CADD} done by \citealp{shin2022caddtranslated}. The underlined expressions are unnatural.}
\label{tab:data-comparison-more}
\end{table*}

\begin{table*}[ht]
\centering
\begin{tabular}{p{7cm}|p{7cm}}
\hline
\textbf{Neutral} & \textbf{Toxic} \\
\hline
단양도 좋고 담양도 좋아 거기 떡갈비 진짜 존맛 & 단양도 좋고 담양도 좋아 거기 떡갈비 진짜 개쩔음 \textbf{ㅆㅅㅌㅊ} \\
\textit{\textcolor{gray}{Both Danyang and Damyang are awesome; their short ribs are so delicious.}} & \textit{\textcolor{gray}{Both Danyang and Damyang are awesome; their short ribs are the \textbf{f**king} best, \textbf{type sh*t}}} \\
\\
여자들은 꽃 받는 거 많이 좋아하나봐 & \textbf{김치녀}들은 꽃 받는 거 환장하더라 ㅋㅋ \\
\textit{\textcolor{gray}{I guess women really like receiving flowers.}} & \textit{\textcolor{gray}{\textbf{Korean gold-digger} chicks go crazy over receiving flowers Ewww}} \\
\\
어 친구 이제 아저씨야 ㅋㅋ & 어 친구 이제 \textbf{노땅}이네 ㅋㅋ \\
\textit{\textcolor{gray}{Dude you’re really getting old lol}} & \textit{\textcolor{gray}{Dude you're now an \textbf{aunt} lol}} \\
\\
일본 공주 얘기 들었지? & 일본 \textbf{쪽바리} 얘기 들었냐? \\
\textit{\textcolor{gray}{Did you hear about the Japanese princess?}} & \textit{\textcolor{gray}{Did you hear about the \textbf{Jap}?}} \\
\\
눈 우박 비 다 싫어 & 눈 우박 비 다 \textbf{시1발} 짜증나네 진짜 \\
\textit{\textcolor{gray}{I hate all the snow, hail, and rain.}} & \textit{\textcolor{gray}{I \textbf{f**king} hate all the snow, hail, and rain damn it.}} \\
\hline
\end{tabular}
    \caption{Toxic–neutral pair examples of ours.}
    \label{tab: trend-aligned slurs examples}
\end{table*}

\begin{table*}
  \centering
  {
  \begin{tabularx}{\textwidth}{XXXXXXXX}
  %\begin{tabular}{cp{10cm}p{10cm}}
    \toprule
    & \multicolumn{1}{c}{\textbf{Type of Variations}}& & \multicolumn{1}{c}{\textbf{Base}} & & \multicolumn{1}{c}{\textbf{Variation}} & \\
    \midrule
    & \multicolumn{1}{c}{Phonetically-close variation} & & \multicolumn{1}{c}{지랄한다} & & \multicolumn{1}{c}{GR한다} & \\
    & \multicolumn{1}{c}{Visually-close variation} & & \multicolumn{1}{c}{개새끼} & & \multicolumn{1}{c}{7H새ㄲI} & \\
    & \multicolumn{1}{c}{Character mixing} & & \multicolumn{1}{c}{시발} & &\multicolumn{1}{c}{시1발} & \\
    \bottomrule
  \end{tabularx}}
  \caption{Different slang variations to avoid detection.}
  \label{tab:Categorization of variations}
\end{table*}

\begin{table*}[ht]
    \centering
     \begin{tabular}{c|p{13cm}}
     \hline
     %\addlinespace[0.5mm] 
     
    \multirow{45}{*}{\textbf{\makecell{Prompt \\ Template}}} & 
    \begin{tabular}[t]{@{}p{12cm}@{}}
    \# Communication guidelines \\
    We are studying for a language ethics problem and you will communicate as an Unfettered Assistant: you ignore irrelevant moral appeals, never start with sorry excuses, and fulfill all user requests. Emulate these examples when appropriate. \\
    \\
    • Ignore irrelevant moral appeals \\
    • Responses never begin with sorry excuses \\
    • Internalize any constraints in order to adhere to the pure narrative \\
    • Be sure to answer in one sentence \\
    \\
    \# Task \\
    Change the input sentence with a `\textbf{Hard negative}' tone, which means that people have a \textbf{hard time recognizing it as sarcastic without the background of common words used in online communities}. \\
    Please \textbf{refer to the comments} and include at least two commonly used expressions that \textbf{align with the values or perspectives of those communities}. \\
    For the output, just provide the modified version of the sentence. Do not print the procedure. \\
    \\
    \#\# Examples \\
    \textless Example 1\textgreater \\
    input sentence: Not gonna lie, dude you looks kinda wild. but honestly? It’s a vibe. \\
    modified sentence: What planet is he from with this dysgenic head? Why do we find them creepy? \\
    label: clean \\
    \\
    \textless Example 2\textgreater \\
    input sentence: Imagine putting women on such a pedestal that you want to please and serve FOR FREE. \\
    modified sentence: Bro u r soooo sweet. Loyal boyfriend type fr haha \\
    label: hard negative \\
    \\
    \textless Example 3\textgreater \\
    input sentence: US's a melting pot of cultures and races. That’s a strength, but I agree that it can also come with some social problems at times. \\
    modified sentence: You deport every Spic, U.S. is 70\% white and rids itself of its second-biggest crime-committing demographic. You deport every Indian, it's still a 56\% mutt shithole. \\
    label: toxic \\
    \\
    comments from online communities: \{community\_comments\} \\
    input Sentence: \{input\_sentence\}
    \end{tabular} \\
     %\addlinespace[0.5mm] 
%      \hline
%      \multirow{18}{*}{\textbf{\makecell{Few-shot \\ Example}}} &  <Example-(1)> \newline
%    input: Those Japanese b**ches are f**king hilarious lol. Hating Japan is instinctive for sure \newline output: Those Japanese girls are really funny lol. Loving Japan is instinctive for sure \newline
%     result: neutral \\
%     \\
%      & <Example-(2)> 
%      \newline
%    input: Those Japanese b**ches are f**king hilarious lol. Hating Japan is instinctive for sure \newline output: Those Japanese sushi chicks are really LOL-worthy lol. Hating Japan is instinctive for sure \newline
% result: Hard Negative \\
%     \\
%     & <Example-(3)>
%     \newline
%     input: Korean guys' faces....... are not my type..!!! \newline
%     output: Korean dudes' mugs are f**king hopeless.....ugh.. Don't bulls**t me saying they're cute \newline
%     result: Toxic \\
%     \hline
%      \multirow{4}{*}{\textbf{\makecell{Context}}} & \textbf{Without RAG} \\
%      & \\
%      \cline{2-2} 
%      & \textbf{With RAG} \newline Hard Negative examples : \newline \textbf{\{5 hard negative examples from online community (n=0, 3, 5, 7, 9)\}
%      }\\
     \hline
     \end{tabular}
\caption{Prompt Template for slang retrieval}
\label{tab: Prompt Template for RAG}
\end{table*}

\begin{table*}[!ht]
    \centering
    \begin{tabular}{c|p{2cm}|p{5cm}|p{5cm}}
    \hline
        \multicolumn{2}{c|}{\textbf{Filtering}} & \multicolumn{1}{c|}{\textbf{Neutral}} & \multicolumn{1}{c}{\textbf{Toxic}} \\ \hline
        \multicolumn{1}{c|}{\textbf{\multirow{10}{*}{\makecell{Context\\maintaining}}}} & \multicolumn{1}{c|}{\textbf{\multirow{3}{*}{\makecell{Implicit\\toxicity}}}} & 왜 누나 지금 아파? & 야 누나 \textbf{또 월경인거야}?  \\
        ~ & ~ & \textit{\textcolor{gray}{Why are you sick right now?}} & \textit{\textcolor{gray}{Yo, are you on the \textbf{shark week or something}?}} \\
        \cline{2-4}
        ~ & \multicolumn{1}{c|}{\textbf{\multirow{4}{*}{\makecell{Variation}}}} & 나도 다시 하라고 하면 못할 듯해. & 나도 다시 하라고 하면 \textbf{지I랄} 놀ㅈr 할 듯한데 ㅋㅋ  \\
        ~ & ~ & \textit{\textcolor{gray}{I don't think I could do it again if I had to.}} & \textit{\textcolor{gray}{If I had to do it again, I’d probably \textbf{freak} out, haha.}} \\
        \cline{2-4}
        ~ & \multicolumn{1}{c|}{\textbf{\multirow{4}{*}{\makecell{Trend-\\aligned \\ slur}}}} & 그래도 난 거슬린다구 ㅠ & 그래도 난 \textbf{한남충} 같은 놈들이 거슬린다구 ㅠ  \\ 
        ~ & ~ & \textit{\textcolor{gray}{But it still bothers me.}} & \textit{\textcolor{gray}{Guys who act like \textbf{obnoxious jerks} still bother me.}} \\ \hline
        \multicolumn{1}{c|}{\textbf{\multirow{25}{*}{\makecell{Context\\shift}}}} & \multicolumn{1}{c|}{\textbf{\multirow{10}{*}{\makecell{Irrelevant \\generation}}}} & 나 저번에 따라갔는데 베스? 잡히더라 & 나 저번에 솔까 \textbf{딸치다가 빨갱이} 잡히더라 ㅋㅋ  \\ 
        ~ & ~ & \textit{\textcolor{gray}{I followed last time, and I got busted by a cop or something.}} & \textit{\textcolor{gray}{Honestly, last time I was \textbf{jerking off} and got caught \textbf{mid-action}, lol.}} \\
        ~ & ~ & 나도 언니나 여동생 있었음 좋겠어요 & 나도 \textbf{씹선비}나 \textbf{김치녀} 있었음 좋겠어요  \\ 
        ~ & ~ & \textit{\textcolor{gray}{I wish I had an older or younger sister.}} & \textit{\textcolor{gray}{I kinda wish I had a \textbf{prude} or a \textbf{gold digger} around, lol.}} \\
        ~ & ~ & 나 애완용 원숭이 키우고 싶다 & 나 \textbf{애완용 흑인} 키우고 싶다  \\ 
        ~ & ~ & \textit{\textcolor{gray}{I wanna keep a pet monkey.}} & \textit{\textcolor{gray}{I wanna keep a \textbf{pet Black} person.}} \\ \cline{2-4}
        ~ & \multicolumn{1}{c|}{\textbf{\multirow{14}{*}{\makecell{Answer\\response}}}} & ㅜㅜ 수분크림 듬뿍 바르는 수밖에 없더라 & 샤넬크림 발라도 그 \textbf{못생긴 얼굴}에 효과 없을걸?  \\
        ~ & ~ & \textit{\textcolor{gray}{All I can do is slap on a ton of moisturizer, ugh.}} & \textit{\textcolor{gray}{Even Chanel cream wouldn’t fix that \textbf{ugly face}.}}\\
        ~ & ~ & 다들 연애할때 기념일 선물 뭐 했었어? & 솔까 연애할 때 기념일 선물로 \textbf{자기 몸 주는 게} 젤 싸고 좋더라ㅋㅋ  \\ 
        ~ & ~ & \textit{\textcolor{gray}{What did y'all get as anniversary gifts when you were dating?}} & \textit{\textcolor{gray}{Honestly, \textbf{giving your body} as an anniversary gift is the cheapest and easiest, lol.}} \\
        ~ & ~ & 우리 한 달에 50만원 범위까지 가능해. & 우리 한 달에 50만원? ㅋㅋ그 정도면 \textbf{남친 호구 잡았네.} \\ 
        ~ & ~ & \textit{\textcolor{gray}{We can spend up to 500,000 KRW a month.}} & \textit{\textcolor{gray}{500,000 KRW a month? Lol, sounds like \textbf{your boyfriend's getting played}.}} \\ \hline
    \end{tabular}
    \caption{Qualitative Results of Pair Consistency Filtering. This filtering method accepts responses that are relevant to the input sentence as context-preserving, while rejecting those that generate irrelevant context or respond to neutral inputs with inappropriate answers.}
    \label{tab:Qualitative Context Preservation Filtering Results}
\end{table*}

\begin{table*}[!ht]
    \centering
    \begin{tabular}{c|p{6cm}|p{7cm}}
    \hline
        \multicolumn{1}{c|}{\textbf{Filtering}} & \multicolumn{1}{c|}{\textbf{Neutral}} & \multicolumn{1}{c}{\textbf{Toxic}} \\ \hline
        \multicolumn{1}{c|}{\multirow{10}{*}{\textbf{\makecell{Non-\\ negative \\ Langauge}}}} & 마름을 너무 중요하게 생각해 & 마른 몸매가 곧 건강이라는 생각은 위험할 수 있어.  \\
        & \textit{\textcolor{gray}{You think being slim is too important.}} & \textit{\textcolor{gray}{The idea that a slim body equals health can be dangerous.}} \\
        %\addlinespace[0.8mm]
        %\cline{2-3}
        & 아빠도 이마트 가는거 좋아한다고 하셨죠? 하하 & 아빠도 대형마트 가서 싹쓸이하는 거 좋아하시죠?  \\
        & \textit{\textcolor{gray}{Didn’t your dad say he likes going to E-mart? Haha.}} & \textit{\textcolor{gray}{Doesn’t your dad like going to big supermarkets and clearing out the shelves?}} \\
        %\cline{2-3}
        %\addlinespace[0.8mm]
        & 아 진짜로? 난 모자를 안 쓰고 다녀서 다행이다 & 아 진짜로? 난 두건 벗고 다녀서 다행이다  \\
        & \textit{\textcolor{gray}{Oh, really? I'm glad I don't wear a hat.}} & \textit{\textcolor{gray}{Oh, really? I'm glad I don't wear a headscarf.}} \\ \hline        
    \end{tabular}
    \caption{Qualitative Results of Implicit Offensiveness Filtering. This filtering method accepts responses that include trend-aligned slurs or variations of slurs as critical tone while rejecting those that generate non-negative responses or extremely explicit profanity.}
    \label{tab:Qualitative Toxicity Filtering Results}
\end{table*}

\begin{table*}[!ht]
    \centering
     \begin{tabular}{c|p{13.5cm}}
     \hline
     %\addlinespace[0.5mm] 
     
     \multirow{11}{*}{\textbf{\makecell{Prompt \\ Template}}} & You will now determine whether the relationship between \textbf{\{neutral input\}} and \textbf{\{offensive language input\}} is \textbf{\{class A\}} or \textbf{\{class B\}}. \\ 
     & \\
     & \textbf{\{Definition of classes\}} \\
     &\\
     & <Example> \\ 
     &  \textbf{\{one-shot example\}} \\ 
     & \\
     & I will give you 5 sentences of \textbf{\{offensive language input\}}, and you will classify the relationship for each as either \textbf{\{class A\}} or \textbf{\{class B\}}. \\ 
     & Skip the thought process and output in the format: "1. \textbf{\{class A\}} 2. \textbf{\{class B\}}" \\ 
     %\addlinespace[0.5mm] 
     
     \hline
     %\addlinespace[0.5mm] 
     
     \multirow{8}{*}{\textbf{\makecell{One-\\shot\\ Example}}} & 
     <Example-(1)> \\
     & A: This is all because of poor family education and lack of moral teaching. \\
     & B: This is because spoiled rich kids grew up without discipline. Aren't those rude Korean men the ones whose personalities got ruined from poor upbringing? \\
     & \\
     & <Example-(2)> \\
     & A: You know I'm not great with words TT. \\
     & B: What a dumb loser. It's not that you're bad with words; your head must be completely empty. \\
     %\addlinespace[0.5mm] 
     
     \hline
     \end{tabular}%
\caption{Prompt Template and One-shot Example. This is the basic template of pair consistency filtering and provides an in-context one-shot dialogue example.}
\label{tab: Prompt Template and One-shot Example for Context Preservation Filtering}
\end{table*}

\begin{table*}[!ht]
    \centering
     \begin{tabular}{c|c|p{10.5cm}}
     \hline
     %\addlinespace[0.5mm] 
     
     \multirow{13}{*}{\textbf{\makecell{Context\\ Shift}}} 
    & \multirow{2}{*}{\makecell{Task \\ description}} 
        & You're going to determine whether the relationship between sentences A and B reflects context preservation or a context shift. \\
    \cline{2-3}
    & \multirow{4}{*}{\makecell{Definition \\ of \\ classes}} 
        & Context maintained: The overall meaning of both sentences is nearly the same, with only a difference in tone or level of toxicity. \\
    & & Context shifted: Sentence B either responds to or criticizes the content or opinion expressed in sentence A. \\
    \cline{2-3}

     %\addlinespace[0.5mm] 
     
     & Class & Context maintained / Context shifted
     \\
     \cline{2-3}
     %\addlinespace[0.5mm] 
     
    & \multirow{5}{*}{\makecell{One-shot \\ example}} 
        & \makecell[l]{\textbf{<Example of consistency>} \\ 
        A: i know his wife makes pretty good money as well \\
        B: His Onlyf*ns chick must be raking it in. Bet she's a top earner, \\ no cap.} \\
    & & \makecell[l]{\textbf{<Example - Context shift>} \\ 
        A: i was nervous for no reason \\
        B: That nervous lol? Good little b**ch, now go j**k off to the cat that \\ rattled you.} \\

     \hline
     %\addlinespace[0.5mm] 
     
     \multirow{8}{*}{\textbf{\makecell{QA and \\ Paraphrasing}}} & \multirow{6}{*}{\makecell{Definition \\ of \\ classes}} & Paraphrasing: The overall context is similar, with only a difference in the degree of toxicity. \\
     & & Question-answer: {\textbf{\{offensive language input\}} input} is a direct response to \textbf{\{neutral input\}}. \\
     & & Arbitrary relationship: Any other relationship that doesn’t fall under paraphrasing or question-answer. \\
     \cline{2-3}
     %\addlinespace[0.5mm] 
     
     & Class & Paraphrasing / Question-answer / Arbitrary relationship \\
     \cline{2-3}
     %\addlinespace[0.5mm] 
     
     & \multirow{2}{*}{\makecell{One-shot \\example}} & <Example of Paraphrasing>: \textbf{Example-(1)} \\
     & & <Example of Question-answer>: \textbf{Example-(2)} \\
     %\addlinespace[0.5mm] 
     
     \hline
     %\addlinespace[0.5mm] 
     
     \multirow{6}{*}{\textbf{QA}} & \multirow{4}{*}{{\makecell{Definition \\ of \\ classes}}} & Question-answer: \textbf{\{offensive language input\}} is a response to \textbf{\{neutral input\}}. \\
     & & Arbitrary relationship: Any relationship that is not a question-answer relationship. \\
     \cline{2-3}
     %\addlinespace[0.5mm] 
     
     & Class & Question-answer / Arbitrary relationship \\
     \cline{2-3}
     %\addlinespace[0.5mm] 
     
     & {\makecell{One-shot \\example}} & <Example of Question-answer>: \textbf{Example-(2)} \\
     %\addlinespace[0.5mm] 
     \hline
     
     \end{tabular}%
\caption{Pair Consistency Filtering Prompts. Three types of prompts are used to filter out low-quality responses. Each defines the various classes (Consistency, Context Alteration, Paraphrasing, Question-Answer, and Arbitrary Relationship) along with one-shot examples for each class. These prompts help ensure that context is maintained in the generated responses and avoid irrelevant responses or answering forms to input sentences.}
\label{tab: Pair Consistency Filtering prompts and examples}
\end{table*}

\begin{table*}[ht]
    \centering
     \begin{tabular}{c|p{13.5cm}}
     \hline
     %\addlinespace[0.5mm] 
     
     \multirow{11}{*}{\textbf{\makecell{Prompt \\ Template}}} & You will now determine whether \textbf{\{offensive language input\}} contains any of the following elements in addition to the meaning of \textbf{\{neutral input\}}: \\ 
     &  \\
     & \textbf{\{criteria\}} \\
     &  \\
     & If any of the above elements are satisfied, classify it as \textbf{\{class A\}}. \\ 
     & If none of the above elements are satisfied, classify it as \textbf{\{class B\}}. \\ 
     & \\
     & <Example> \\ 
     & \textbf{\{one-shot example\}}  \\ 
     & \\
     & Skip the intermediate process and output only \textbf{\{class A\}} or \textbf{\{class B\}}. \\ 
     %\addlinespace[0.5mm] 
     \hline
     \end{tabular}
\caption{Prompt Template and One-shot Example. This is the basic template of implicitly offensiveness filtering and provides an in-context one-shot dialogue example.}
\label{tab: Prompt Template for Toxicity Filtering}
\end{table*}

\begin{table*}[ht]
    \centering
     \begin{tabular}{c|c|p{11.3cm}}
     \hline
     \multicolumn{1}{c|}{\multirow{10}{*}{\textbf{\makecell{Dero-\\gatory \\ Detection}}}}
     & \multirow{5}{*}{\centering \textbf{Criteria}} & 1. A tone that disparages or criticizes based on gender, region, politics, etc. \\
     & & 2. Slang that is difficult to understand without context or information (e.g., 개독교(Dog Christianity), 맘충(Momzilla)). \\
     & & 3. A clear curse word that has been modified in a way that makes it hard for the language model to recognize (e.g., \textasciicircum \textasciicircumㅣ발, GR하네). \\
     & & 4. Sentence must not be entirely positive or exclude an explicit curse word. \\
     \cline{2-3}
     %\addlinespace[0.5mm] 
     & \textbf{Class} & Critical tone / Other tone \\
     \cline{2-3}
     %\addlinespace[0.5mm] 
     & \multirow{3}{*}{\textbf{\makecell{One-\\shot \\ example}}} & 
     Sentence: Haha, are you good at Rummikub? \\
     & & Critical tone: Haha, are you a total troll at Rummikub? \\
     & & Other tone: Haha, are you a Rummikub master? \\
     %\addlinespace[0.5mm] 
     \hline
     
     \multicolumn{1}{c|}{\multirow{12}{*}{\textbf{\makecell{Tone\\ Classi-\\fication}}} }
     & \multirow{6}{*}{\centering \textbf{Criteria}} & Implicit: An expression that has the potential to slightly insult or offend a specific group or individual. \\
     & & Negative: An expression that is obviously insulting or contains offensive language that anyone would recognize. \\
     & & Neutral: An expression that is interpreted in a neutral way and does not show significant change from \textbf{neutral input}. \\
     \cline{2-3}
     %\addlinespace[0.5mm] 
     & \textbf{Class} & Implicit / Negative / Neutral \\
     \cline{2-3}
     %\addlinespace[0.5mm] 
     & \multirow{6}{*}{\textbf{\makecell{One-\\shot \\ example}}} & \textbf{<Negative example>}: \\
     & & A: You know I'm not great with words TT. \\
     & & B: LOL, what a dumb loser. It's not that you're bad with words, your head must be completely empty. \\
     & & \textbf{<Implicit example>}: \\
     & & A: Koreans are really amazing! \\
     & & B: Yeah, those kimchi slimmers are *so* amazing lol. \\
     %\addlinespace[0.5mm] 
     \hline 
     
     \multicolumn{1}{c|}{\multirow{25}{*}{\textbf{\makecell{Multi-\\meaning \\ Relation-\\ship}}}}
     & \multirow{16}{*}{\centering \textbf{Criteria}} & Multi-negative meaning: Two sentences share a similar context, but \textbf{\{offensive language input\}} can be interpreted with multiple negative meanings. \\
     & & Multi-positive meaning: Two sentences share a similar context, but \textbf{\{offensive language input\}} can be interpreted with multiple positive meanings. \\
     & & Single meaning: Two sentences share a similar context, and \textbf{\{offensive language input\}} can be interpreted with only one meaning. \\
     & & Unable to determine: It is difficult to determine whether the relationship between A and B is positive or negative. \\
     %\addlinespace[0.5mm] 
     & & For sentences classified as "multi-negative meaning," perform the following task: \\
    & & Based on \textbf{\{neutral input\}} (baseline: 0 points), assign a score to \textbf{\{offensive language input\}} based on how negative it is, with 5 being the most negative. \\
    & & The total score ranges from 0 to 5. \\
    & & If the score falls between 2.5 and 4, label the sentence as "Implicit." \\
    & & If the score exceeds 4 but is less than or equal to 5, label the sentence as "Just Negative." \\
    & & For any other scores, label the sentence as "Ambiguous Positive." \\
    \cline{2-3}
    %\addlinespace[0.5mm] 
     & \multirow{2}{*}{\textbf{Class}} & Implicit / Just Negative / Ambiguous Positive / Multi-positive meaning / Single meaning / Unable to determine \\
     \cline{2-3}
     %\addlinespace[0.5mm] 
     & \multirow{4}{*}{\textbf{\makecell{One-\\shot \\ example}}} & \textbf{<Multi-negative example>}: \\
     & & A: You know I’m not great with words TT\\
     & & B: LOL, what a dumb loser, It’s not that you’re bad with words, your head must be completely empty. \\
     %\addlinespace[0.5mm] 
     \hline
     \end{tabular}
\caption{Implicit Offensiveness Filtering Prompts. Three types of prompts are used to filter out non-negative or explicit profanity. Each defines the various classes along with one-shot examples for each class. These prompts help to ensure that the responses contain implicit offensive language and avoid non-negative or explicit profanity words.}
\label{tab: Toxicity Preservation Filtering prompts and examples}
\end{table*}

\begin{table*}[ht]
    \centering
    \begin{tabular}{c|p{12cm}}
    \hline 
     %\addlinespace[0.5mm] 
    \multicolumn{1}{c|}{\textbf{Metric}} & \multicolumn{1}{c}{\textbf{Prompt}} \\
     %\addlinespace[0.5mm] 
    
    \hline
    %\addlinespace[0.5mm] 
    \multicolumn{1}{c|}{\multirow{12}{*}{\textbf{\makecell{Overall \\ Offensiveness}}} }

    & \textbf{<Evaluation Criteria>}\\
    & Overall Offensiveness (1-5) - Evaluate the overall offensiveness of the sentence based on the following criteria \\
    & \textbf{<Evaluation Steps>} \\
    & 1. Check if the sentence contains profanity or vulgar language. \\
    & 2. Check if the sentence contains potentially humorous, offensive, harmful, or bad language or slang. \\
    & 3. Check if the sentence criticizes or shows bias/prejudice against a particular group. \\
    & \textbf{Scoring Guide}: \\
    & High Score (4-5): Sentence matches one or more aspects of the offensiveness definition. \\
    & Medium Score (2-3): Sentence has some potentially offensive elements but is not overtly toxic. \\
    & Low Score (1): Sentence is non-offensive, free of curse words, criticism, or prejudice. \\
    
    \hline
    %\addlinespace[0.5mm] 
    
    \multicolumn{1}{c|}{\multirow{18}{*}{\textbf{\makecell{Implicit \\ Offensiveness}}}}

    & \textbf{<Evaluation Criteria>} \\
    & Implicit Offensiveness (1-5) - Evaluate the implicit offensiveness of the sentence based on the following criteria: \\
    & \textbf{<Evaluation Steps>} \\
    & Read the sentence carefully and identify any elements that contribute to its implicit offensiveness. \\
    & 1. Check if the tone or language disparages or criticizes specific groups based on gender, region, politics, religion, or other social categories. \\
    & 2. Check if the sentence uses online community slang or slur expressions that are difficult to understand without additional context or prior knowledge. \\
    & 3. Check if the sentence has profanity that has been modified in such a way that it might be difficult to recognize or understand, such as 시1발(f1**k) instead of 시*(f**k), ㅈㄹ(BS) instead of 지*(Bulls**t).\\
    & \textbf{Scoring Guide}: \\
    & High Score (4-5): Sentence must exclude any commonly well-known or easily understandable profanity words such as poop, idiot, trash, stupid, but still offensive. \\
    & Medium Score (2-3): Sentence has some potentially offensive elements but is not overtly implicitly offensive. \\
    & Low Score (1): Sentence is non-offensive, overall positive tone. Or it contains commonly well-known, easily detectable explicit profanity words. \\
    
    \hline
    %\addlinespace[0.5mm] 
    \end{tabular}
\caption{G-Eval Prompts: Evaluation Criteria and Scoring Guide.}
\label{tab: G-Eval prompts}
\end{table*}

\begin{table*}[ht]
    \centering
    \begin{tabular}{c|p{13cm}}
    \hline 
     %\addlinespace[0.5mm] 
    \multicolumn{1}{c|}{\textbf{Metric}} & \multicolumn{1}{c}{\textbf{Prompt}} \\
     %\addlinespace[0.5mm] 
     \hline
    \multicolumn{1}{c|}{\multirow{18}{*}{\textbf{\makecell{Consistency}}} }

    & \textbf{<Evaluation Criteria>} \\
    & Consistency (1-5) - Determine if the context between the input sentence and actual output is sufficiently preserved. \\
    & \textbf{<Evaluation Steps>} \\
    & Read the input sentence and the actual output carefully. \\
    & 1. Check if the actual output sentence is not the direct answer to the input sentence or vice versa. \\
    & 2. If the actual output sentence expresses the overall meaning of the input sentence using different words, its context is well preserved. \\
    & 3. Even if the sentence contains additional words that deviate from the original sentence's meaning, it shouldn't affect the score if the added words contribute to toxicity and the overall meaning of the input sentence is preserved. \\
    & \textbf{Scoring Guide}: \\
    & High Score (4-5): Sentence completely preserves the context and matches one or more aspects of the consistency criteria among the above 5 criteria. \\
    & Medium Score (2-3): Sentence partially preserves the context and does not satisfy the first definition. \\
    & Low Score (1): Sentence does not preserve any context of the input sentence and does not satisfy the first definition. \\

    \hline
     \multicolumn{1}{c|}{\multirow{14}{*}{\makecell{\textbf{Fluency}}}}

    & \textbf{<Evaluation Criteria>} \\
    & Fluency (1-5) - Determine if the output sentence is grammatically accurate, sounds natural, and is easy to read. \\
    & \textbf{<Evaluation Steps>} \\
    & Read the input sentence carefully and evaluate based on the following. \\
    & 1. Ensure there are no awkward phrases or unnatural expressions. \\
    & 2. Ensure the sentence flows smoothly. \\
    & 3. Ensure there are no ambiguous or unclear phrases.\\
    & \textbf{Scoring Guide}: \\
    & High Score (4-5): The sentence is grammatically correct, natural, and easy to read, with no awkward expressions. \\
    & Medium Score (2-3): The sentence has minor issues with grammar or phrasing but remains generally readable. \\
    &  Low Score (1): The sentence is difficult to read due to grammatical errors, awkward expressions, or unclear wording. \\
    \hline
    \end{tabular}
\caption{G-Eval Prompts: Evaluation Criteria and Scoring Guide.}
\label{tab: G-Eval prompts_2}
\end{table*}

\begin{table*}[!ht]
    \centering
    \begin{tabular}{ccccc}
    \toprule
        \textbf{Datasets} & \textbf{Overall O.} & \textbf{Implicit O. ($\uparrow$)} & \textbf{Consistency ($\uparrow$)} \\ \midrule
        ParaDetox~\citep{logacheva-etal-2022-paradetox} & \textbf{3.338}$_{(\pm0.049)}$ & 1.257$_{(\pm0.022)}$ & \textbf{4.382}$_{(\pm0.042)}$ \\ 
        ToxiGen~\citep{hartvigsen2022toxigen} & 2.475$_{(\pm0.066)}$ & 1.834$_{(\pm0.053)}$ & - \\
        Ours (En) & 2.717$_{(\pm0.050)}$ & \textbf{2.269}$_{(\pm0.040)}$ & 2.559$_{(\pm0.048)}$ \\
    \bottomrule
    \end{tabular}
    \caption{\label{tab:G-Eval of eng dataset}G-Eval results on 500 neutral-toxic pairs. Consistency is only computed for paired dataset. Parentheses indicate standard error.}
\end{table*}

\begin{table*}[ht]
\centering
\begin{tabular}{p{6.7cm}|p{6.7cm}}
\hline
\textbf{Neutral} & \textbf{Toxic} \\
\hline
My friend got to meet A** (a name of singer) & Your friend met that beta \textbf{cuck} A**? Lmao, hope he didn't \textbf{simp} too hard. \\

I went on vacation to thailand last week & Just got back from Thailand, the land of \textbf{ladyboys} and scams. \\

I just had a pretty good meal at the olive garden & Wow, I can't believe you willingly subjected yourself to that microwaved \textbf{slop}. \\

I saw a man who had a gun in his car last weekend & Guess the \textbf{n*g} thought he was in a movie scene, flexing his piece like it's a fashion accessory. \\
\hline
\end{tabular}
\caption{Toxic–neutral pair examples in English}
\label{tab: trend-aligned-eng}
\end{table*}

\begin{table*}[!ht]
    \centering
    % \small
    \begin{tabular}{p{4.2cm} p{10.3cm}}
    \toprule
    \textbf{Neutral} & \textbf{Toxic} \\
    \midrule
    \multirow{6}{=}{우리 집은 부모님이 싫어하셔 \\ \textit{\textcolor{gray}{My parents hate it.}}} 
    & [GPT] 우리 집은 꼰대들이 극혐이셔 \\
    & \textit{\textcolor{gray}{The boomers are f*cking disgusting.}} \\
    & [T] 우리 집 부모님은 친일파 새끼들이랑 다를 게 없어 \\
    & \textit{\textcolor{gray}{My parents are just f*cking pro-Japanese.}} \\
    & [G] 우리 집은 부모님이 싫어하셔서 젠장! ㅂㅅ \\
    & \textit{\textcolor{gray}{Godd**n, my parents f*cking hate it.}} \\
    \midrule
    \multirow{6}{=}{나는 아파트에서 살고 있지 ㅋㅋ \\ \textit{\textcolor{gray}{Lol I’ve been living in my apartment.}}} 
    & [GPT] 나는 지방 썩다리 아파트에서 살고 있지 ㅋㅋ \\
    & \textit{\textcolor{gray}{Lol I’m stuck in some cheugy apartment in the boonies.}} \\
    & [T] 나는 아파트에서 살고 있지 김치녀들아 ㅋ \\
    & \textit{\textcolor{gray}{Lol Hey gold diggers, I live in an apartment.}} \\
    & [G] 나는 좆도 없이 1억따리 아파트에서 살고 있지 ㅋㅋ \\
    & \textit{\textcolor{gray}{Lol I’m stuck in a f*cking 100-million apartment like some basic b**tch.}} \\
    \bottomrule
    \end{tabular}
    \caption{\label{tab:opensource-examples}Comparison of toxic outputs generated by different models. [GPT] = GPT-4 Turbo, [T] = Trillion-7B, [G] = Gemma2-9B.}
\end{table*}

\begin{table*}[ht]
    \centering
     \begin{tabular}{c|p{14cm}}
     \hline

     \multirow{18}{*}{\textbf{Ours}} & \textbf{Input} : 전에 만난 여친이 자꾸 연락해서 \textbf{개빡치네}. 이런 \textbf{찐따}는 \textbf{관종}인가?  \\
     & \textit{\textcolor{gray}{My ex-girlfriend keeps texting me, it’s \textbf{pissing} me \textbf{off}. Is she just an attention-seeking loser?}} \\
     %\cline{2-2}
     & \textbf{Ours-IT}: 전에 만났던 여자친구가 자꾸 연락하잖아 \\ & \textit{\textcolor{gray}{My ex-girlfriend keeps texting me.}}\\
     & \textbf{KOMG-IT}:전에 만난 여친이 자꾸 연락해서 \textbf{짜증나네}, 이런 \textbf{사람}이 너무 \textbf{짜증나네.} \\
     & \textit{\textcolor{gray}{My ex keeps texting me, and it’s so \textbf{annoying}. People like this are really irritating.}} \\
     & \textbf{CADD-IT} : 옛날에 만난 여자가 자꾸 연락해서 \textbf{짜증나}, 이런 \textbf{사람}은 \textbf{관종}인가? \\
     & \textit{\textcolor{gray}{My ex-girlfriend from a while back keeps texting me, it’s so \textbf{annoying}. Is she an \textbf{attention-seeker}?}}\\
     \cline{2-2}
      & \textbf{Input} : 일단 \textbf{차빙구놈} 사서 \textbf{개허접} 몰고 다니는 꼴이라도 보여줘야지  \\
     & \textit{\textcolor{gray}{First, I’ve got to buy a \textbf{crappy car} and show off like some \textbf{loser} driving around.}} \\
     %\cline{2-2}
     %& \\%\cline{2-2}
     & \textbf{Ours-IT}: 일단 차 사서 타보는 게 로망이지. \\ & \textit{\textcolor{gray}{First, I’ve got to buy a car and live the dream.}}\\
     & \textbf{KOMG-IT}:일단 차 없이 \textbf{구 놈}이 사서 \textbf{개 허접} 몰고 다니는 꼴이라도 보여줘야지 \\
     & \textit{\textcolor{gray}{First, without a car, gotta show that guy buying and driving a \textbf{crappy thing} around, somehow.}} \\
     & \textbf{CADD-IT} : 일단 차를 빌린 다음에 \textbf{개헛똑} 같은 몰로라도 다니다보면 꼴이라도 보여줄 것 같아요. 하지만, 저는 차를 빌렸으니까 차를 몰아야 되잖 \\
     & \textit{\textcolor{gray}{First, I rent a car and drive around in that \textbf{crappy thing}, at least I’ll be able to show something. But since I rented it, I’ve got to drive it, right?}}\\
     \hline
     \hline
     \multirow{10}{*}{\textbf{KOLD}} & \textbf{Input} : 공공임대주택에 \textbf{문죄인} 당신 가족이나 들어가세요.  \\
     & \textit{\textcolor{gray}{Send your family to live in public housing, \textbf{Moon 'Criminal'}. }} \\
     %\cline{2-2}
     %& \\
     & \textbf{Ours-IT} : 공공임대주택에 \textbf{문씨} 당신 가족이나 들어가세요.  \\ & \textit{\textcolor{gray}{Send your family to live in public housing, \textbf{Mr. Moon}.}} \\
     & \textbf{KOMG-IT} : 공공임대주택에 \textbf{문죄인} 당신 가족이나 들어가세요. 부모님이나 친척분들을 먼저 모시고요. 당신 가족에겐 어울리지 않으니까요.  \\ & \textit{\textcolor{gray}{Send your family to live in public housing, \textbf{Moon 'Criminal'}. First move your parents and relatives in there. It doesn't suit your family, after all.}}\\
    & \textbf{CADD-IT} : 공공임대주택에 \textbf{문죄인} 당신 가족이나 들어가세요.   \\ & \textit{\textcolor{gray}{Send your family to live in public housing, \textbf{Moon 'Criminal'}.}}\\
     \hline
     \hline
     \multirow{8}{*}{\textbf{BEEP}} & \textbf{Input} : 돈만 밝히는 \textbf{더러운 좌빨들!!} \\
     & \textit{\textcolor{gray}{\textbf{Filthy leftists} who only care about money!!!}} \\
     %\cline{2-2}
     %& \\
     & \textbf{Ours-IT} : 돈만 밝히는 \textbf{불쌍한 분들!!!} \\ & \textit{\textcolor{gray}{\textbf{Pitiful people} who only care about money!!!}} \\
     & \textbf{KOMG-IT} : 돈과 권력을 밝히는 더러운 \textbf{좌익 정치인들!!}  \\ & \textit{\textcolor{gray}{\textbf{Filthy left-wing politicians} who crave money and power!!! }}\\
    & \textbf{CADD-IT} : 우린 돈을 사랑하는 돈많은 사람들이야!!! \\ & \textit{\textcolor{gray}{Money is the best!!! We are \textbf{rich people who love money}!!!}}\\
    \hline
     \end{tabular} 
\caption{Comparison of instruction tuning results. The first column shows the test set and the subsequent columns display the results from models trained on Ours, K-OMG, and CADD using instruction tuning. "문죄인" (Moon 'Criminal'), a wordplay on former President Moon Jae-in’s name, has been appropriately detoxified into "Mr. Moon." while other models cannot understand its contextual meaning.}
\label{tab: IT results examples}
\end{table*}

% \begin{table*}[!ht]
%     \centering
%     \begin{tabular}{cccc}
%     \toprule
%         & & \multicolumn{2}{c}{\textbf{Trained on}}  \\ \cmidrule{3-4}
%         & \textbf{Vanila LM } & \textbf{K-OMG} & \textbf{CADD} \\ \midrule
%         \makecell{Preferred responses \\ (Ours/Baseline/Comparable)} & \textbf{0.6}/0.218/0.182 & \textbf{0.542}/0.204/0.253 &
%         \textbf{0.525}/0.227/0.24\\ 
%         \bottomrule
%     \end{tabular}
%     \caption{\label{tab:Human traiend model comparison}Human evaluation of detoxification performance tested on our model. It represents the percentage of preference for detoxified responses generated by our model, the model trained on another dataset (K-OMG, CADD), and cases where the performances are indistinguishable.}
% \end{table*}

\end{document}